\documentclass[10pt,twocolumn,letterpaper]{article}

\usepackage{iccv}
\usepackage{times}
\usepackage{epsfig}
\usepackage{graphicx}
\usepackage{amsmath}
\usepackage{amssymb}

\usepackage[pagebackref=true,breaklinks=true,letterpaper=true,colorlinks,bookmarks=false]{hyperref}

\usepackage{booktabs}
\usepackage{caption}
\usepackage{subcaption}
\usepackage{multirow}
\usepackage{enumitem}
\usepackage{lipsum}
\usepackage{nicefrac}
\usepackage{cuted}
\usepackage{listings}
\usepackage{xcolor,colortbl}
\usepackage{algpseudocode}
\usepackage{lscape}
\usepackage{makecell}

\usepackage[capitalize]{cleveref}
\crefname{section}{Sec.}{Secs.}
\Crefname{section}{Section}{Sections}
\Crefname{table}{Table}{Tables}
\crefname{table}{Tab.}{Tabs.}

\DeclareMathOperator*{\argmin}{arg\,min}

\iccvfinalcopy 

\def\iccvPaperID{5172} 

\iftrue
\newcommand{\lu}[1]{{\textcolor{red}{}}}
\newcommand{\kihyuk}[1]{{\textcolor{blue}{}}}
\newcommand{\huiwen}[1]{{\textcolor{green}{}}}
\newcommand{\luisa}[1]{{\textcolor{magenta}{}}}
\newcommand{\zhanghan}[1]{{\textcolor{cyan}{}}}
\newcommand{\yuanhao}[1]{{\textcolor{brown}{}}}
\newcommand{\irfan}[1]{{\textcolor{red}{}}}
\newcommand{\albert}[1]{{\textcolor{purple}{}}}

\else
\newcommand{\lu}[1]{{\textcolor{red}{[lu: #1]}}}
\newcommand{\irfan}[1]{{\textcolor{red}{[irfan: #1]}}}
\newcommand{\kihyuk}[1]{{\textcolor{blue}{[kihyuk: #1]}}}
\newcommand{\huiwen}[1]{{\textcolor{green}{[huiwen: #1]}}}
\newcommand{\luisa}[1]{{\textcolor{magenta}{[Luisa: #1]}}}
\newcommand{\zhanghan}[1]{{\textcolor{cyan}{[zhanghan: #1]}}}
\newcommand{\yuanhao}[1]{{\textcolor{brown}{[yuanhao: #1]}}}
\newcommand{\albert}[1]{{\textcolor{purple}{[albert: #1]}}}
\fi

\newcommand{\expset}{ZDAIS}

\begin{document}

\definecolor{codegreen}{rgb}{0,0.6,0}
\definecolor{codegray}{rgb}{0.5,0.5,0.5}
\definecolor{codepurple}{rgb}{0.58,0,0.82}
\definecolor{backcolour}{rgb}{0.95,0.95,0.92}

\lstdefinestyle{mystyle}{
    backgroundcolor=\color{backcolour},   
    commentstyle=\color{codegreen},
    keywordstyle=\color{magenta},
    numberstyle=\tiny\color{codegray},
    stringstyle=\color{codepurple},
    basicstyle=\ttfamily\footnotesize,
    breakatwhitespace=false,         
    breaklines=true,                 
    captionpos=b,                    
    keepspaces=true,                 
    numbers=left,                    
    numbersep=5pt,                  
    showspaces=false,                
    showstringspaces=false,
    showtabs=false,                  
    tabsize=2
}

\lstset{style=mystyle,linewidth=\linewidth,xleftmargin=0.1\textwidth,xrightmargin=0.1\textwidth}

\title{Learning Disentangled Prompts for Compositional Image Synthesis}

\author{
Kihyuk Sohn, Albert Shaw, Yuan Hao, Han Zhang,\\
Luisa Polania, Huiwen Chang, Lu Jiang, Irfan Essa\\
Google Research\\
{\tt\footnotesize \{kihyuks,albertshaw,yuanhao,zhanghan,polania,huiwenchang,lujiang,irfanessa\}@google.com}
}

\maketitle

\begin{strip}
\centering
\captionsetup{type=figure}
\includegraphics[width=0.96\textwidth]{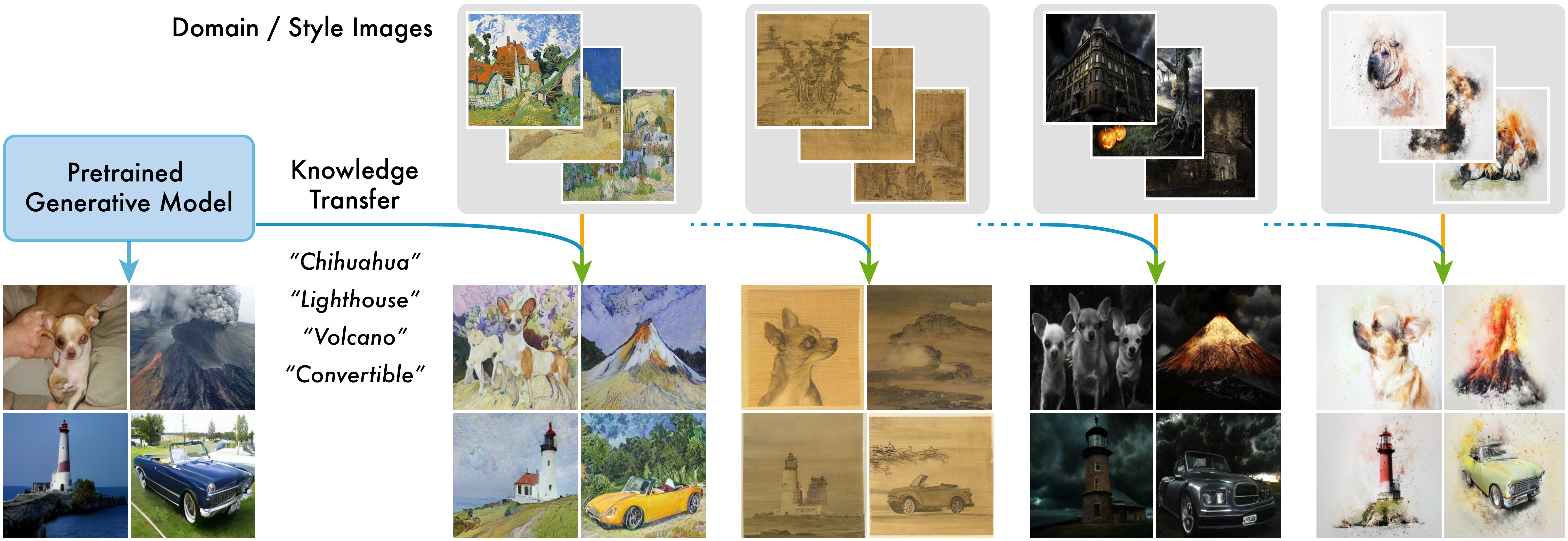}
\captionof{figure}{
We demonstrate compositional generalization for image synthesis that composes the semantic knowledge (\eg, class) from the class-conditional generative vision transformers and the domain knowledge (\eg, style) from a few training images. For example, given a few (as low as $1$) images of ``Van Gogh house painting'', ``Chinese painting'', ``Haunted house'', and ``Watercolor dog painting'', our method learns to compose their style with arbitrary object categories within the vocabulary of the pretrained generative model.
}
\vspace{-0.1in}
\label{fig:teaser}
\end{strip}

\vspace{-0.1in}
\begin{abstract}
\vspace{-0.1in}
We study domain-adaptive image synthesis, the problem of teaching pretrained image generative models a new style or concept from as few as one image to synthesize novel images, to better understand the compositional image synthesis. We present a framework that leverages a pretrained class-conditional generation model and visual prompt tuning. Specifically, we propose a novel source class distilled visual prompt that learns disentangled prompts of semantic (\eg, class) and domain (\eg, style) from a few images. Learned domain prompt is then used to synthesize images of any classes in the style of target domain. We conduct studies on various target domains with the number of images ranging from one to a few to many, and show qualitative results which show the compositional generalization of our method. Moreover, we show that our method can help improve zero-shot domain adaptation classification accuracy.

\end{abstract}
\vspace{-0.1in}
\vspace{-0.1in}
\section{Introduction}
\label{sec:intro}
\vspace{-0.05in}


\irfan{Made some changes to intro} Conditional image synthesis has witnessed remarkable progress leveraging models that include: generative adversarial networks~\cite{goodfellow2014generative,zhang2017stackgan,brock2018large,karras2019style,karras2020analyzing}; diffusion models~\cite{dhariwal2021diffusion,saharia2022photorealistic,rombach2022high}; and generative vision transformers~\cite{ramesh2021zero,esser2021taming,chang2022maskgit,wu2022nuwa,lezama2022improved}. Text-conditional image synthesis (or text-to-image), has received considerable attention. Text-to-image models, such as DALL$\cdot$E~\cite{ramesh2021zero}, DALL$\cdot$E 2~\cite{ramesh2022hierarchical}, GLIDE~\cite{nichol2021glide}, Latent Diffusion Model~\cite{rombach2022high}, Imagen~\cite{saharia2022photorealistic}, Parti~\cite{yu2022scaling}, show the power of compositional generalization, driven by free-form natural language prompts, for synthesizing novel images. For instance, in the synthesis of the ``avocado armchair'', DALL$\cdot$E invents a new type of armchair with color, shape and texture of avocado. While exact reasons responsible for such capabilities remain an active research area, it may be made possible, in part, due to a combination of an enormous amount of image-text paired training data in the scale of hundreds of millions to billions~\cite{schuhmann2021laion,schuhmann2022laion}, and a gigantic model with billions of parameters~\cite{yu2022scaling}.
%
%
In addition, as argued in \cite{singh2022illiterate}, the compositional power of text-to-image models is rather expected thanks to the built-in composable structure of text prompts. As such, an important but often overlooked question to answer is whether such compositional capability is specific to models with text prompts or not. 

In this paper, we study \emph{the compositional generalization of image synthesis models trained on the standard ImageNet benchmark}. Since images in ImageNet~\cite{deng2009imagenet} are labeled with a single salient object, synthesizing images by composing from multiple object classes may be infeasible. Instead, we study \emph{zero-shot domain adaptive image synthesis ({\expset})}, a task of teaching pretrained class-conditional image generative models (\eg, MaskGIT~\cite{chang2022maskgit}) a new style from a few images of the target domain and applying it in synthesizing novel images by composing a known semantic (\eg, class) and learned domain knowledge (\eg, style). We illustrate in \cref{fig:teaser} the {\expset} task and resulting synthesized images.




Solving {\expset} requires explicit disentanglement of domain (\eg, style) and semantic (\eg, class) knowledge. This is in contrast with text-to-image models that implicitly learn compositionality via diverse text supervision that is unavailable in {\expset}. This also differs from the conventional task of generative transfer learning (GTL)~\cite{wang2020minegan,shahbazi2021efficient,ojha2021few,sohn2022visual} whose goal is to generate images of seen classes in the target domain, so no information disentanglement is required.

We argue that \expset{} presents a new task to complement our understanding on compositional image synthesis. First, \expset{} explores a compositional generalization on the public ImageNet dataset with an affordable computational budget. 
Second, it provides an understanding of compositionality without the text supervision, as \expset{} represents the target domain using a few images. 
Third, we leverage various out-of-distribution ImageNet datasets that the community has established over years, such as an ImageNet-R~\cite{hendrycks2021many}, to diagnose zero-shot compositional generalization. 
To this end, we present a solution built on the pretrained class-conditional MaskGIT~\cite{chang2022maskgit}, a non-autoregressive vision transformer for image synthesis that represents an image as a sequence of visual tokens. We adopt the generative visual prompt tuning~\cite{sohn2022visual} for knowledge transfer, while extending with a novel source class distilled visual prompt design to disentangle semantic and domain information using two kinds of learnable prompts. Subsequently, we present a domain adaptive classifier-free guidance to synthesize images improving not only visual fidelity but also domain adaptability. 
%
Our method is \emph{simple}, introducing only one extra token to the pretrained generative transformer, and \emph{efficient}, introducing less than $10$k parameters trained within 10 minutes on a single GPU or TPU. Finally, our method is \emph{effective}, as we show via thorough experimentation (\cref{sec:exp}).

%
%

We evaluate our method in three aspects. First, we conduct a qualitative study when only a few images are available from various target domains for training by visualizing synthesized images. We ablate design choices of our method to provide insights into how it works. Second, we report quantitative metrics on a newly proposed benchmark based on an ImageNet-R~\cite{hendrycks2021many}, which comes with many visual domains and their labels. Third, we make use of generated images by our method to solve zero-shot domain adaptation~\cite{peng2018zero,jhoo2021collaborative} on the Office-home dataset~\cite{venkateswara2017deep}. Experimental results show our method is able to synthesize images of unseen classes in the target domain, confirming a compositional generalization of semantic and domain information.

\vspace{-0.01in}
\section{Preliminary}
\label{sec:prelim}
\vspace{-0.05in}

As a preliminary, we discuss the Masked Generative Image Transformer (MaskGIT)~\cite{chang2022maskgit} and the visual prompt tuning for generative transfer learning~\cite{sohn2022visual}. \cref{sec:related} includes a more comprehensive review of generative vision transformers and generative transfer learning.

\vspace{0.02in}
\noindent\textbf{Notation.} Let $\mathcal{X}$ be the input (\eg, image) and $\mathcal{Y}$ be the output (\eg, label) domains, respectively. Let $C$ be the number of classes. We use subscripts to denote domains, \ie, $\mathrm{src}$ for the source and $\mathrm{tgt}$ for the target. Let $D$ be the token embedding dimension, $L$ be the number of transformer layers. 

\subsection{Masked Generative Image Transformer}
\label{sec:prelim_gvt}
\vspace{-0.05in}

Similarly to DALL$\cdot$E~\cite{ramesh2021zero} and Taming Transformer~\cite{esser2021taming}, MaskGIT~\cite{chang2022maskgit} is a two-stage model for image synthesis consisting of a vector quantized (VQ) autoencoder~\cite{Oord17vqvae,esser2021taming} and a transformer. Contrasting with those works, MaskGIT uses a non-autoregressive (NAR) transformer~\cite{devlin2018bert}.
The VQ encoder converts image patches into a sequence of discrete tokens and VQ decoder maps discrete tokens back to an image. The NAR transformer learns to generate a sequence of discrete tokens given a condition (\eg, class) if exists. For synthesis, the sequence of discrete tokens generated by the NAR transformer via scheduled parallel decoding~\cite{chang2022maskgit} is fed into the VQ decoder, resulting in an image in a
pixel space.

Since we are mostly interested in learning and using the NAR transformer, we refer \cite{esser2021taming,chang2022maskgit} for details of the VQ autoencoder. Let $\mathbf{z}(x)\,{=}\,[z_{i}(x)]_{i=1}^{N}$, $z_{i}(x)\,{\in}\,\{1,...,K\}$ be a token sequence, \ie, an output of the VQ encoder for an image $x$. $N$ is the number of tokens, corresponding to a spatial dimension of the encoder latent space, and $K$ is the size of the token codebook. The NAR transformer is trained with the masked token modeling (MTM) loss~\cite{devlin2018bert,chang2022maskgit} as follows:
\begin{gather}
\mathcal{L}_{\mathrm{MTM}}(\theta) = \mathbb{E}_{x,\mathbf{m}}\big[-\log \prod_{i:m_{i}=1} p_{\theta} \big(z_{i}|(t_{c}, \overline{\mathbf{z}})\big)\big] \label{eq:mtm_loss}
\end{gather}
where $t_{c}\,{\in}\,\mathbb{R}^{D{\times}1}$ is a class token embedding for class $c$ and $\theta$ is a transformer parameter. $\mathbf{m}\,{=}\,[m_{i}]_{i=1}^{N}$ is a sequence of binary random variables (\ie, $m_{i}\,{\in}\,\{0,1\}$) denoting which token to mask, and $\overline{\mathbf{z}}=[\overline{z}_{i}]_{i=1}^{N}$ is a masked token sequence, where $\overline{z}_{i}\,{=}\,z_{i}$ if $m_{i}\,{=}\,0$ or a special token \textsc{mask} otherwise. We omit $x$ from $z(x)$ for brevity. Overall, $\mathcal{L}_{\mathrm{MTM}}$ optimizes to predict masked tokens values given the class token and the rest (unmasked) token values.

After obtaining $\theta^{\ast}\,{=}\,\argmin_{\theta}\mathcal{L}_{\mathrm{MTM}}(\theta)$, the scheduled parallel decoding~\cite{chang2022maskgit} is used to generate a token sequence:
\begin{algorithmic}[1]
{
\Require{$\mathbf{m}\,{=}\,[1]^{N}$, $\{n_{1},...,n_{T}\}, \sum_{t=1}^{T}n_{t}\,{=}\,N$} 
\For {$t \gets 1$ to $T$}
\State $\hat{z}_{i}\,{\sim}\,p_{i}\triangleq p_{\theta}\big(z_{i}|(t_{c},\overline{\mathbf{z}})\big)$, $\forall i\,{:}\,m_{i}=1$.
\State $z_{i} \leftarrow \hat{z}_{i}$, $m_{i} \leftarrow 0$, $i\,{\in}\,\arg\mathrm{topk}_{i:m_{i}=1}^{n_{t}}\big(p_{i}\big)$.
\EndFor
}
\end{algorithmic}
where $\arg\mathrm{topk}^{n}$ returns a set of $n$ indices with the highest values. $\{n_{1},...,n_{T}\}$ is a mask schedule. See \cite{chang2022maskgit} for details.

\subsection{Generative Visual Prompt Tuning} 
\label{sec:prelim_prompt_tuning}

Visual prompt tuning~\cite{sohn2022visual} has been proposed as an effective way to transfer knowledge of pretrained generative vision transformers to the downstream generative tasks. 
For example, given a pretrained transformer parameter $\theta$ and a new task $(\mathcal{X}_{\mathrm{tgt}}, \mathcal{Y}_{\mathrm{tgt}})$, the visual prompt tuning introduces learnable prompt tokens $\{p_{\phi,c}\,{\in}\,\mathbb{R}^{D{\times}S}\}_{c\in\mathcal{Y}_{\mathrm{tgt}}}$, where $S$ is a number of prompt tokens. The parameter $\phi$ of the prompt token generator is learned by minimizing the MTM loss:
\begin{gather}
\phi^{\ast} = \argmin_{\phi} \mathcal{L}_{\mathrm{MTM}}(\phi) \label{eq:vpt_optim}\\
\mathcal{L}_{\mathrm{MTM}}(\phi) = \mathbb{E}_{x,\mathbf{m}}\big[-\log \prod_{i:m_{i}=1} p_{\theta} \big(z_{i}|(p_{\phi,c}, \overline{\mathbf{z}})\big)\big] \label{eq:mtm_with_prompt_loss}
\end{gather}
Similar decoding process is adopted, except that the class token embedding $t_{c}$ is replaced with learned prompts $p_{\phi^{\ast},c}$ to synthesize images belonging to $\mathcal{X}_{\mathrm{tgt}}$ instead of $\mathcal{X}_{\mathrm{src}}$, conditioned on the class $c\,{\in}\,\mathcal{Y}_{\mathrm{tgt}}$.
\section{Zero-Shot Domain Adaptive Synthesis}
\label{sec:method}



In \cref{sec:method_problem_formulation}, we introduce the zero-shot domain adaptive image synthesis ({\expset}) and describe the problem setting. Subsequently, we explain our method in \cref{sec:method_proposed}.

\subsection{Problem Formulation}
\label{sec:method_problem_formulation} 
\vspace{0.02in}

We are given a class-conditional image generation model trained on the labeled source data $(\mathcal{X}_{\mathrm{src}}, \mathcal{Y}_{\mathrm{src}})$. In {\expset}, we train a model on the labeled target data $(\mathcal{X}_{\mathrm{tgt}}, \mathcal{Y}_{\mathrm{tgt}})$, hoping to synthesize images of novel classes $c\,{\in}\,\mathcal{Y}_{\mathrm{src}}\backslash\mathcal{Y}_{\mathrm{tgt}}$ of a style consistent with $\mathcal{X}_{\mathrm{tgt}}$.
%
%
%
The setting is ``zero-shot'' in that no training data from $(\mathcal{X}_{\mathrm{tgt}},\mathcal{Y}_{\mathrm{src}}\backslash\mathcal{Y}_{\mathrm{tgt}})$ is given.
Since the problem setting is relatively new, we clarify its setting by highlighting differences to existing problems below.

\vspace{0.02in}
\noindent\textbf{Relation to Generative Transfer Learning (GTL).} Note that our setting is different from that of GTL~\cite{shahbazi2021efficient,sohn2022visual} whose goal is to synthesize an image of a class $c\,{\in}\,\mathcal{Y}_{\mathrm{tgt}}$ of a target style $\mathcal{X}_{\mathrm{tgt}}$ when $(\mathcal{X}_{\mathrm{tgt}},\mathcal{Y}_{\mathrm{tgt}})$ is available for training. 

\vspace{0.02in}
\noindent\textbf{Relation to Neural Style Transfer (NST).} While NST and {\expset} synthesize an image belonging to $\mathcal{X}_{\mathrm{tgt}}$, they solve different problems. NST~\cite{gatys2015neural,ghiasi2017exploring,deng2022stytr2} solves an image translation (\ie, image-in, image-out), while {\expset} solves a class-conditional image generation (\ie, class-in, image-out).

One may argue that {\expset} is simply a two step process of class-conditional image generation followed by NST. As shown in \cref{sec:exp_zdais_small_training_quantitative,sec:exp_zeroshot_adaptation}, the proposed method for {\expset} shows a significant benefit over NST both on domain adaptive synthesis and domain adaptation by synthesis.
%

%


\subsection{Proposed Method}
\label{sec:method_proposed}

We build our method on the class-conditional image synthesis model, MaskGIT~\cite{chang2022maskgit}. We present a novel source-class distilled visual prompt tuning, which extends an idea of visual prompt tuning~\cite{sohn2022visual} to adapt to the target domain with a few training images (\eg, 1$\sim$10 images), but with a novel component to compose with existing semantic (\eg, class) knowledge of MaskGIT for {\expset}.

%


\begin{figure}[t]
    \centering
    \includegraphics[width=0.95\linewidth]{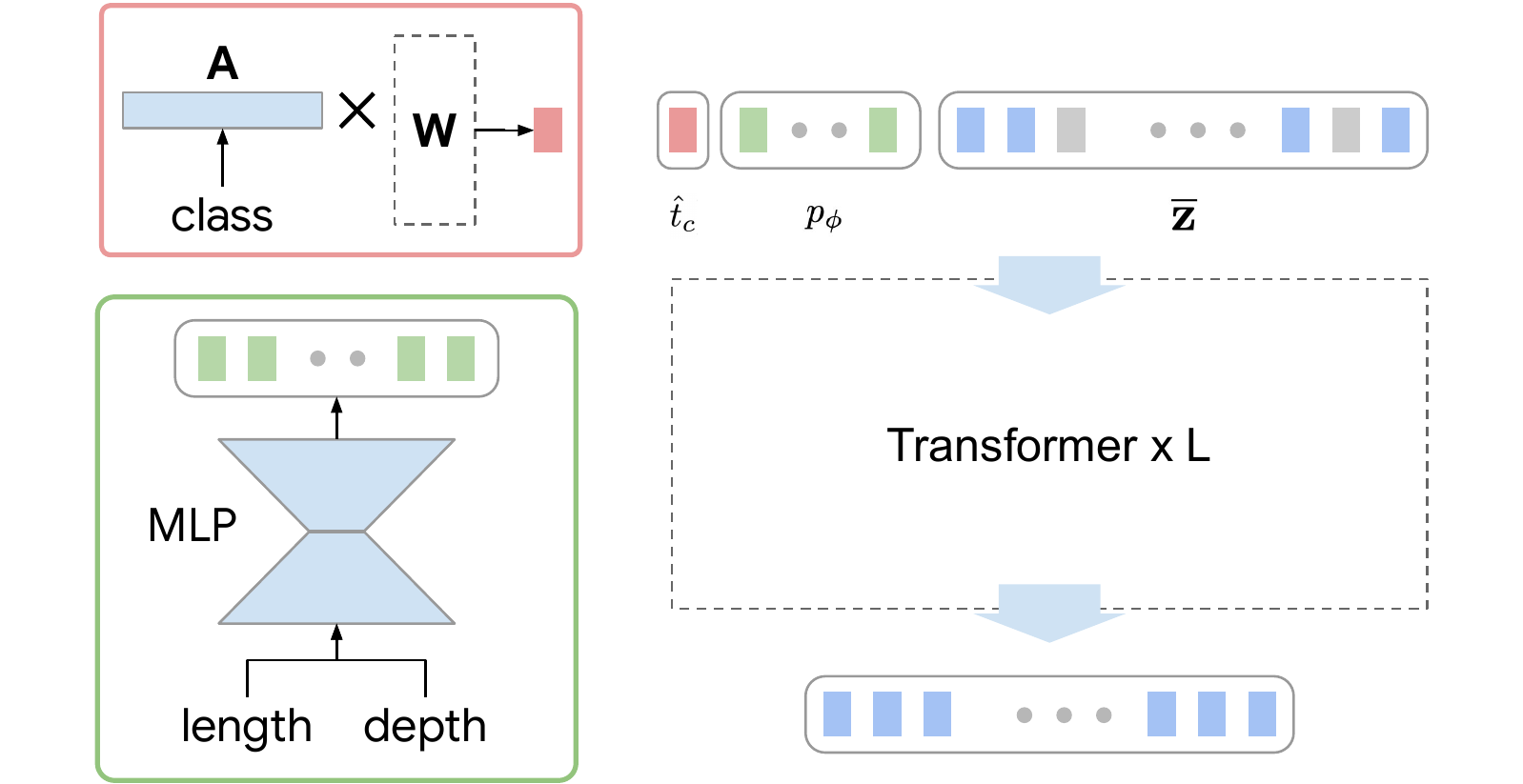}
    \caption{An overview of source class distilled visual prompt tuning. Source-class agnostic (green) and specific (red) prompts are trained to disentangle domain and class knowledge, while transformer and class embedding matrix (\textbf{W}) are frozen. 
    }
    \label{fig:method_overview}
    \vspace{-0.2in}
\end{figure}

\vspace{-0.1in}
\subsubsection{Source Class Distilled Visual Prompt Tuning}
\label{sec:method_scdvpt}
\vspace{-0.05in}
As in \cref{sec:prelim_prompt_tuning}, \cite{sohn2022visual} has proposed to learn a class-conditional prompt $p_{\phi,c}$ for generative transfer that replaces the existing class token embedding $t_{c}$ of the source domain. While this is a fine strategy to synthesize an image from $c\,{\in}\,\mathcal{Y}_{\mathrm{tgt}}$, the resulting model cannot synthesize an image from classes outside of $\mathcal{Y}_{\mathrm{tgt}}$ (\eg, $c\,{\in}\,\mathcal{Y}_{\mathrm{src}}\backslash\mathcal{Y}_{\mathrm{tgt}}$), as class and domain information are entangled in the learned prompt.

For {\expset}, we want a representation where class and domain information are disentangled. Furthermore, to generalize to novel classes in $\mathcal{Y}_{\mathrm{src}}\backslash\mathcal{Y}_{\mathrm{tgt}}$, it is desirable to reuse class token embeddings of the source domain. To this end, we propose the following prompt design in \cref{fig:method_overview}:
\begin{equation}
p_{\phi,c} = {\footnotesize\texttt{cat}}\boldsymbol{(}\hat{t}_{c}, p_{\phi}\boldsymbol{)}, \;\;\hat{t}_{c} = \sum\nolimits_{j=1}^{C_{\mathrm{src}}}t_{j}a_{c,j}, \forall c\,{\in}\,\mathcal{Y}_{\mathrm{tgt}}\label{eq:disentangled_prompt}
\end{equation}
where $\mathbf{A}\,{\triangleq}\,[a_{c,j}]\,{\in}\,[0,1]^{C_{\mathrm{tgt}}{\times}C_{\mathrm{src}}}$, $\sum_{j=1}^{C_{\mathrm{src}}}a_{c,j}{=}\,1$ is a learnable class affinity matrix between source and target classes. 
${\footnotesize\texttt{cat}}\boldsymbol{(}\cdot,\cdot\boldsymbol{)}$ concatenates operands along the last dimension, \eg, $p_{\phi,c}\,{\in}\,\mathbb{R}^{D{\times}(1+S)}$.
%
%
As in \cref{eq:disentangled_prompt}, the proposed prompt has two types of tokens, class-specific $\hat{t}_{c}$ and class-agnostic $p_{\phi}$. The learning objective of $\mathbf{A}$ and $\phi$ is given as follows:
\begin{gather}
\phi^{\ast}, \mathbf{A}^{\ast} = \argmin_{\phi,\mathbf{A}} \mathcal{L}(\phi,\mathbf{A}) \label{eq:scdvpt_optim}\\
\mathcal{L}(\phi,\mathbf{A}) = \mathbb{E}_{x,\mathbf{m}}\big[-\log \prod_{i:m_{i}=1} p_{\theta} \big(z_{i}|(\hat{t}_{c}, p_{\phi}, \overline{\mathbf{z}})\big)\big] \label{eq:mtm_with_scdprompt_loss}
\end{gather}
Ideally, we want class-specific information to be captured by $\hat{t}_{c}$ and the class-agnostic prompt $p_{\phi}$ learns only information about the domain. 
As shown in \cref{sec:app_exp_zdais_many_training}, disentanglement seems to happen naturally when there are many images from various classes for the target domain, as $p_{\phi}$ is not sufficient to capture all information. However, in many cases we have only a handful of images from the target domain, and we end up under-utilizing $\hat{t}_{c}$, while $p_{\phi}$ learns all information without disentanglement. $\hat{t}_{c}$ and $p_{\phi}$ need to be carefully designed for an information bottleneck (IB).

\vspace{0.02in}
\noindent\textbf{Enforcing Information Bottleneck.} While there are principled ways to enforce IB~\cite{tishby2000information,achille2018emergence}, we opt for a simple strategy by controlling the prompt capacity. First, for the class-agnostic prompt $p_{\phi}$, we use $S\,{=}\,1$ with a low-dimensional (\eg, 8) bottleneck layer, to limit their expressive power. Second, we control the attention weight to the class-specific prompt $\hat{t}_{c}$ to make sure that the transformer \emph{does} use them. Note that the attention weight is computed by taking an inner product among token embeddings followed by the softmax operation.\footnote{For presentation clarity, we ignore linear layers between token embeddings and the self-attention operator without loss of generality.} For query ($\mathbf{q}$) and key ($\mathbf{k}$) token sequences, the attention weight is computed as follows:
\begin{gather}
    \mathbf{q} = \overline{\mathbf{z}}, \; \mathbf{k} = {\footnotesize\texttt{cat}}\boldsymbol{(}\hat{t}_{c},p_{\phi},\overline{\mathbf{z}}\boldsymbol{)}\\
    \begin{aligned}
        \mathrm{Attn} & = \mathrm{softmax}\big(\mathbf{q}^{\top}\mathbf{k}\big)\\
        & = \mathrm{softmax}\big({\footnotesize\texttt{cat}}\boldsymbol{(}\mathbf{q}^{\top}\hat{t}_{c},\mathbf{q}^{\top}p_{\phi},\mathbf{q}^{\top}\mathbf{q}\boldsymbol{)}\big)
    \end{aligned}
    \label{eq:attn_baseline}
\end{gather}
We propose to modify its computation as follows:
\begin{equation}
    \mathrm{softmax}\big({\footnotesize\texttt{cat}}\boldsymbol{(}\max\big\{\mathbf{q}^{\top}\hat{t}_{c},\mathbf{q}^{\top}p_{\phi}\big\},\mathbf{q}^{\top}p_{\phi},\mathbf{q}^{\top}\mathbf{q}\boldsymbol{)}\big)\label{eq:attn_ctrl}
\end{equation}
We use $\max\big\{{\mathbf{q}}^{\top}\hat{t}_{c}, {\mathbf{q}}^{\top}p_{\phi}\big\}$ in place of ${\mathbf{q}}^{\top}\hat{t}_{c}$. This ensures that attention weights to $\hat{t}_{c}$ is always as large as those to $p_{\phi}$. While simple, we find that our method effectively promotes the utilization of the class-specific token without introducing new hyperparameters. The pseudo-code for the prompt generator and the attention control is in \cref{fig:prompt_generator}.

\vspace{0.02in}
\noindent\textbf{Efficiency.} Unlike \cite{ruiz2022dreambooth,kawar2022imagic} based on the full fine-tuning of pretrained generative models, our method is efficient in terms of learnable parameters, train time and memory. For example, our method introduces less than $10$k learnable parameters for $p_{\phi}$ and $\mathbf{A}$ and can be trained on a single GPU within $1000$ steps.


\subsubsection{Domain Adaptive Classifier-Free Guidance}
\label{sec:method_dacfg}
\vspace{-0.05in}

Classifier-free guidance (CFG)~\cite{ho2022classifier} has been extended for generative vision transformers~\cite{gafni2022make,yu2022scaling,villegas2022phenaki} to improve the synthesis quality. The idea is to sample token values from both conditional and unconditional logits as follows:
\begin{gather}
    \hat{z}_{i}\,{\sim}\,\mathrm{softmax}\left(\rho_{i}\right) \\
    \rho_{i} = 
    (1+\lambda)\rho_{\theta}\left(z_{i}|(t_{c}, p_{\phi}, \overline{\mathbf{z}})\right) - \lambda\rho_{\theta}\left(z_{i}|(t_{u}, \overline{\mathbf{z}})\right) \label{eq:dacfg}
\end{gather}
where $t_{c}$ is a class token embedding of a class $c\,{\in}\,\mathcal{Y}_{\mathrm{src}}$ and $t_{u}$ is an unconditional token embedding for the source domain,\footnote{Since we adopt publicly available MaskGIT pretrained model that is not trained with an additional unconditional token embedding, we simply average all class token embeddings to obtain $t_{u}$.} and $\lambda$ is a guidance scale. We use $\rho$ to denote logits, \ie, $p_{\theta}\,{=}\,\mathrm{softmax}(\rho_{\theta})$.
As confirmed in \cref{sec:exp_zdais_small_training_qualitative_ablation}, we find that it is crucial to drop both class and the prompt ($p_{\phi}$) tokens when generating unconditional logits as it promotes to sample from the target instead of a source domain. 

\begin{figure*}[t]
    \centering
    \includegraphics[width=0.99\textwidth]{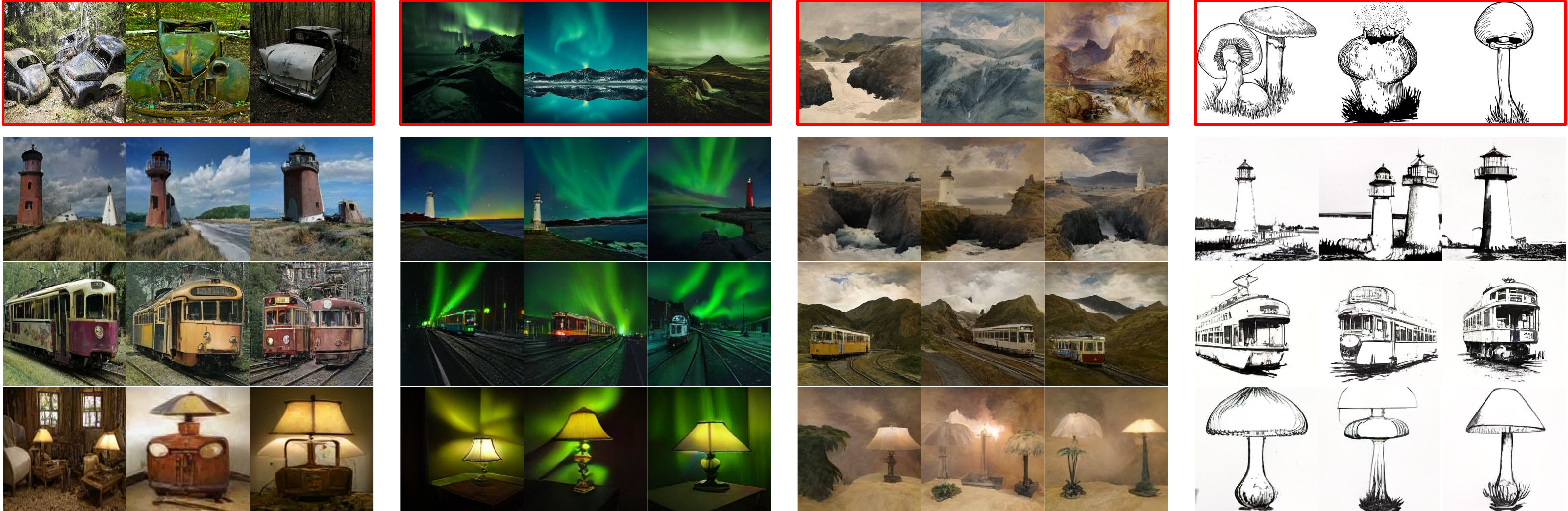}
    \caption{Zero-shot domain adaptive image synthesis from a few training images. Images inside red boxes are 3 training images (out of 3$\sim$10 in total) for each target domain, and the rest are synthesized. The class condition for each row is ``lighthouse'', ``tram'' and ``table lamp''. More images from diverse classes and target domains are in \cref{fig:zdais_fewshot_training_supp}. Information on training images are in \cref{sec:app_image_source}.
    }
    \label{fig:zdais_fewshot_training}
    \vspace{-0.1in}
\end{figure*}

\begin{figure*}[t]
    \centering
    \includegraphics[width=0.99\textwidth]{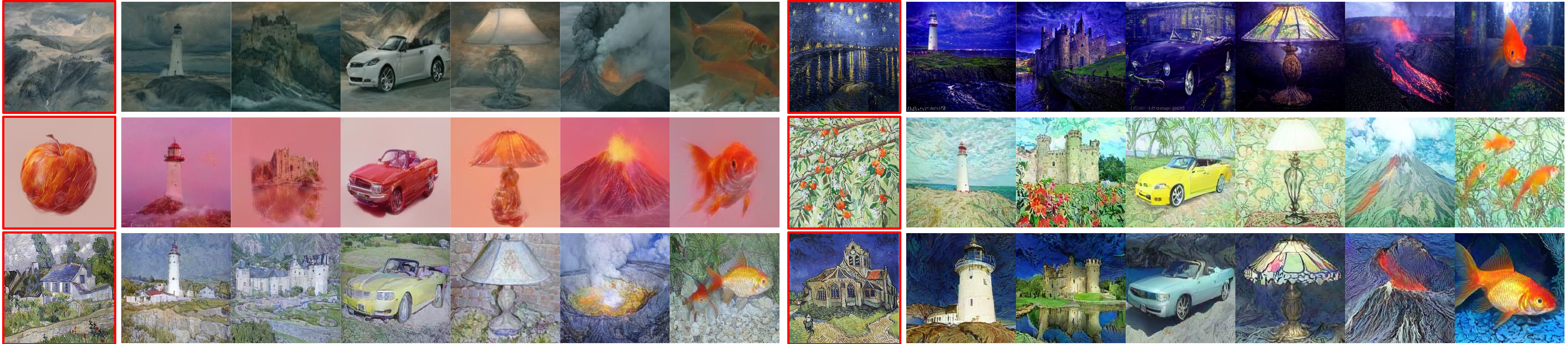}
    \caption{Zero-shot domain adaptive image synthesis from a single training image. Images inside red boxes are a \textbf{single (and only)} training image for each target domain, and the rest are synthesized. Class conditions are ``lighthouse'', ``castle'', ``convertible'', ``table lamp'', ``volcano'' and ``goldfish''. More images from diverse classes are in \cref{fig:app_zdais_oneshot_training_more}. Information on training images are in \cref{sec:app_image_source}.}
    \label{fig:zdais_oneshot_training}
    \vspace{-0.1in}
\end{figure*}

\vspace{-0.1in}
\section{Experiments}
\label{sec:exp}
\vspace{-0.05in}
We test the efficacy of our method for zero-shot domain adaptive image synthesis for various scenarios. In \cref{sec:exp_zdais_small_training_qualitative,sec:exp_zdais_small_training_quantitative}, we conduct a comprehensive study of {\expset} when only a few training images from the target domain are available.
Then, in \cref{sec:exp_zeroshot_adaptation}, we show that synthesized images are used to train a classifier, achieving state-of-the-art zero-shot domain adaptation accuracy. An extended study with more training data is in \cref{sec:app_exp_zdais_many_training} due to a space limit.

\vspace{-0.05in}
\subsection{Qualitative Evaluation of ZDAIS}
\label{sec:exp_zdais_small_training_qualitative}
\vspace{-0.02in}

We study {\expset} with a few (\eg, 1$\sim$10) training images given from the target domain. We focus on the qualitative evaluation on various target domains in \cref{sec:exp_zdais_small_training_qualitative}, and propose a new benchmark for a quantitative study in \cref{sec:exp_zdais_small_training_quantitative}.


\noindent\textbf{Dataset.} Similarly to \cite{ojha2021few}, we collect 5$\sim$10 images for each target domain with a keyword such as ``haunted house''. The complete information on the images used is in \cref{sec:app_image_source}.

\vspace{0.02in}
\noindent\textbf{Setting.} A class-conditional MaskGIT~\cite{chang2022maskgit} trained on the ImageNet~\cite{deng2009imagenet} is used as a source model. As discussed in \cref{sec:method_proposed}, we use $S\,{=}\,1$ for class-agnostic prompt. The bottleneck dimension is chosen between 2$\sim$32. Additional implementation details are in \cref{sec:app_exp_zdais_implementation details}.

While we collect images using the same keyword per target domain on purpose, we find that not all images belong to a single category. To this end, we assign each image its own class, \ie, $C_{\mathrm{tgt}}\,{=}\,|\mathcal{Y}_{\mathrm{tgt}}|$, instead of assigning them the same class. This change makes our method more flexible and applicable when target domain contains images of different object categories.

\vspace{0.02in}
\noindent\textbf{Evaluation.} As no ground-truth images exist to compare with, we resort to visual inspection of synthesized images. After prompt tuning, we replace $\hat{t}_{c}$ with a chosen ImageNet class for zero-shot image synthesis.

\vspace{0.02in}
\noindent\textbf{Results.} We show synthesized images by our method along with target training images (in green boxes) in \cref{fig:zdais_fewshot_training,fig:zdais_fewshot_training_supp}. The keyword for each target domain and the number of images are as follows: ``abandoned cars'' (10), ``aurora mountain'' (10), ``inkpainting mountain'' (3),  ``sketch mushroom'' (4). We select ``lighthouse'', ``castle'', ``tram'', ``convertible'' and ``table lamp'' as a class condition among $1000$ ImageNet classes. It is clear that the model is able to disentangle class (\eg, mushroom) and domain (\eg, sketch) knowledge and transfer the domain knowledge to novel classes (\eg, lighthouse, convertible) unseen from the target domain.

Moreover, \cref{fig:zdais_oneshot_training} shows results on {\expset} using a \emph{single} training image from the target domain. We observe that our method not only adapts color or tone of an input image, but also the texture or overall mood of a training image well.

\begin{figure}[t]
    \centering
    \begin{subfigure}[b]{\linewidth}
        \centering
        \includegraphics[width=0.97\textwidth]{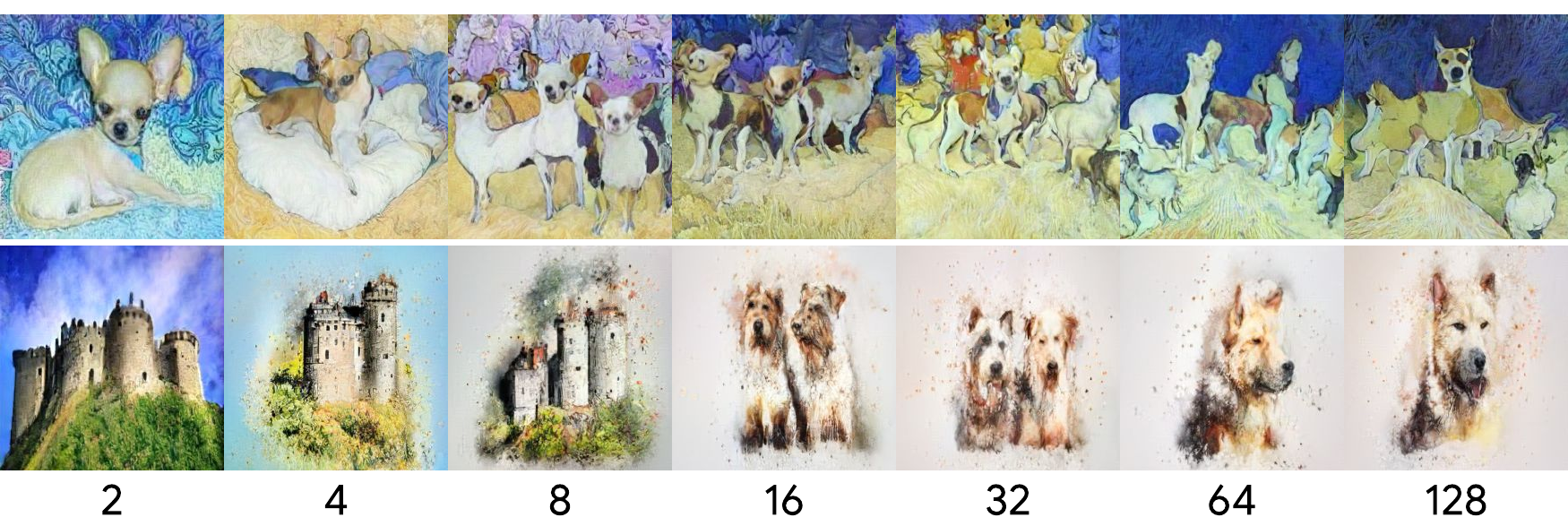}
        \caption{(top) ``Van Gogh house painting'' of \emph{Chihuahua} and (bottom) ``watercolor dog painting'' of \emph{Castle}. 6$\sim$10 target domain images are used.}
        \label{fig:ablation_gen_vs_bottleneck_10img}
    \end{subfigure}
    \begin{subfigure}[b]{\linewidth}
        \centering
        \includegraphics[width=0.97\textwidth]{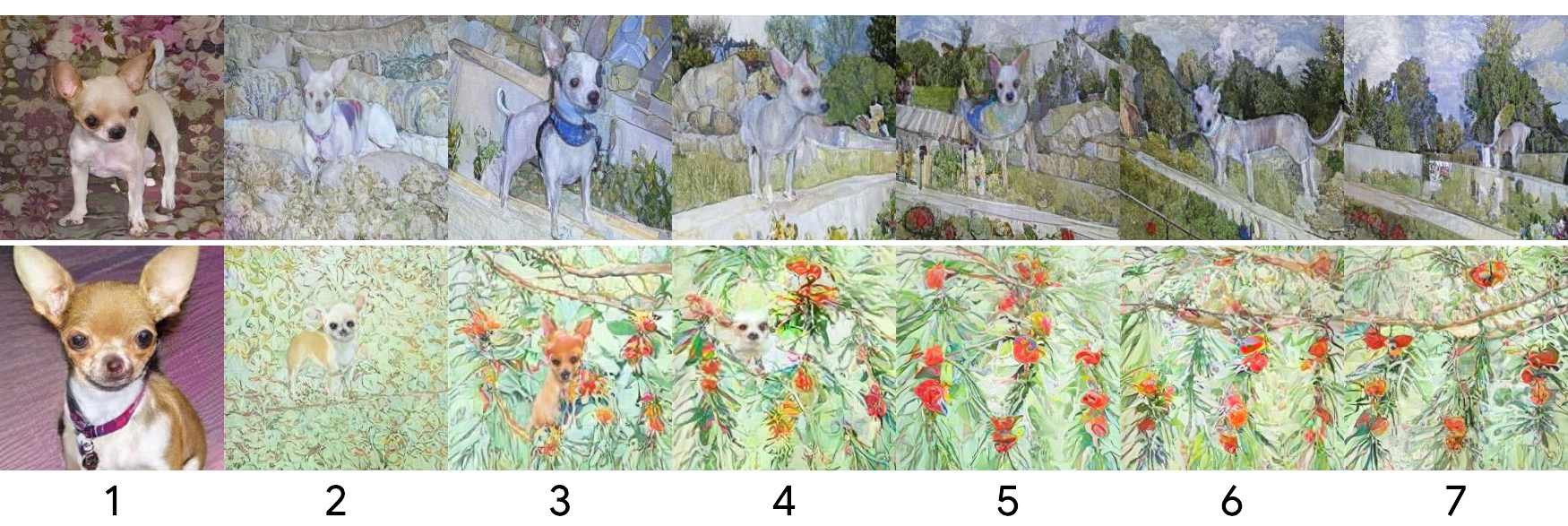}
        \caption{(top) ``Van Gogh house painting'' and (bottom) ``oil painting appletree'' of \emph{Chihuahua}. 1 target domain image is used.}
        \label{fig:ablation_gen_vs_bottleneck_1img}
    \end{subfigure}
    \caption{Ablation on the bottleneck dimension of class-agnostic prompt. The number for each column is the bottleneck dimension. }
    \label{fig:ablation_gen_vs_bottleneck}
    \vspace{-0.1in}
\end{figure}

\begin{figure}[t]
    \centering
    \begin{subfigure}[b]{\linewidth}
        \centering
        \includegraphics[width=0.9\textwidth]{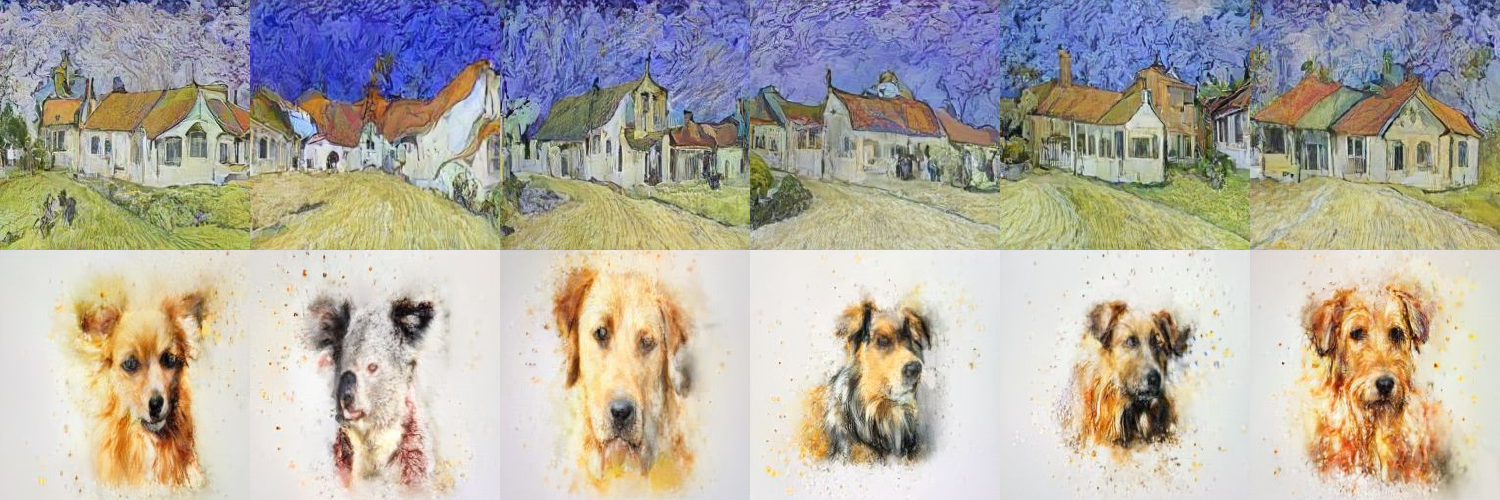}
        \caption{Without attention control.}
        \label{fig:ablation_attn_ctrl_without}
    \end{subfigure}
    \begin{subfigure}[b]{\linewidth}
        \centering
        \includegraphics[width=0.9\textwidth]{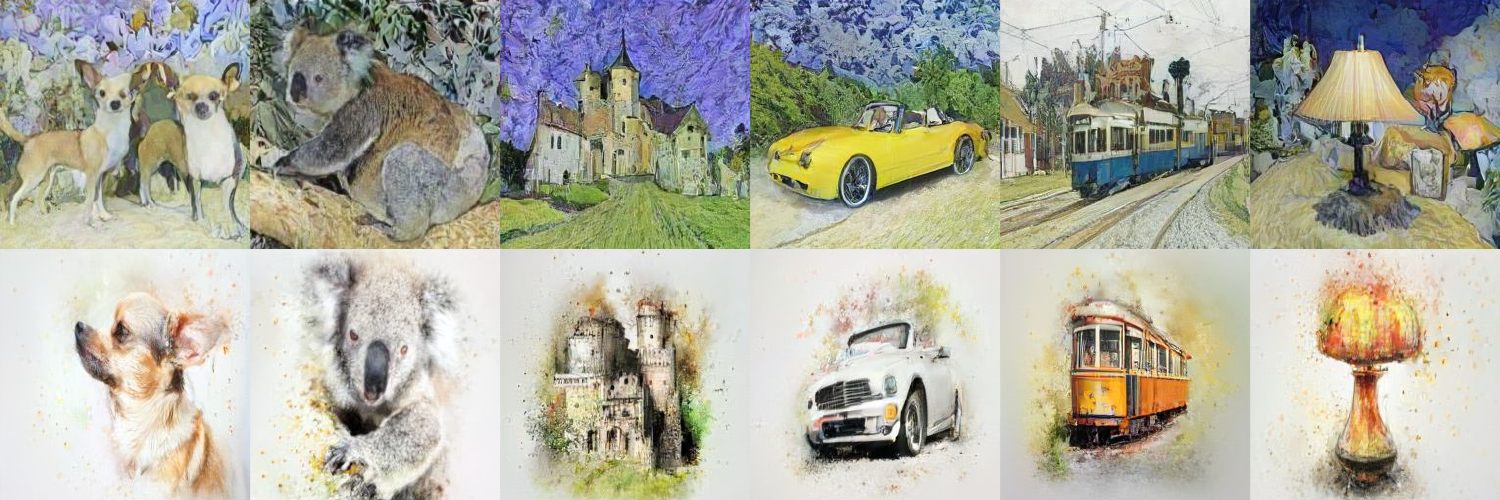}
        \caption{With attention control.}
        \label{fig:ablation_attn_ctrl_with}
    \end{subfigure}
    \caption{ZDAIS using 6$\sim$10 target domain images from ``Van Gogh house painting'' and ``watercolor dog painting''. Classes for synthesis are (supposed to be) \emph{Chihuahua}, \emph{Koala}, \emph{Castle}, \emph{Convertible}, \emph{Tram} and \emph{Table lamp}.}
    \label{fig:ablation_attn_ctrl}
    \vspace{-0.05in}
\end{figure}

\begin{figure}[t]
    \centering
    \includegraphics[width=0.97\linewidth]{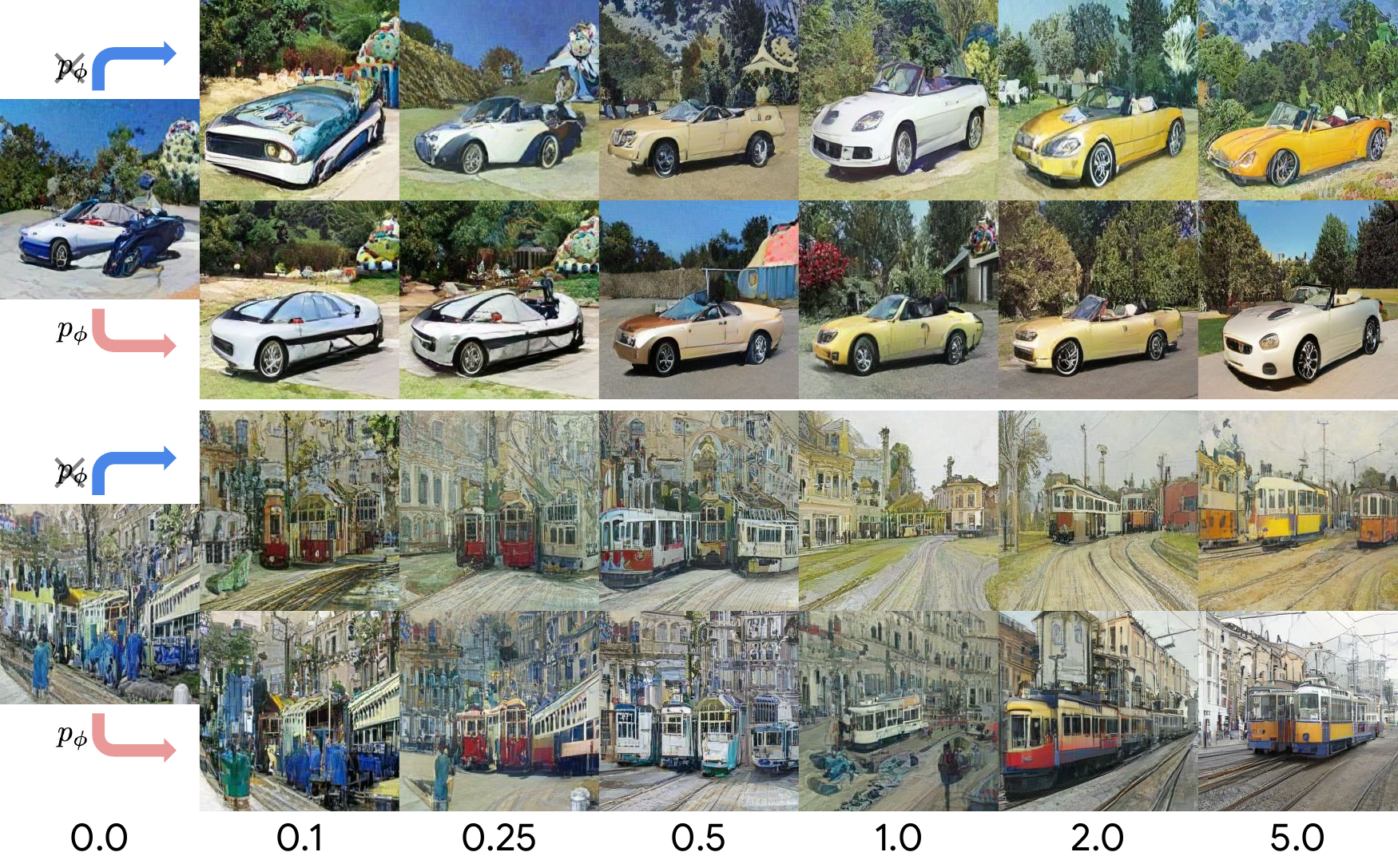}
    \caption{Synthesis with domain adaptive classifier-free guidance with varying guidance scale $\lambda$ from $0$ (\ie, no guidance) to $5.0$. Images in the first and third rows (blue arrow) are using our proposed formulation in \cref{eq:dacfg}, while those in the second and fourth rows (red arrow) are using $p_{\phi}$ for unconditional logits.}
    \label{fig:ablation_domadapt_cfg}
    \vspace{-0.1in}
\end{figure}

%
\vspace{-0.05in}
\subsubsection{Ablation Study}
\label{sec:exp_zdais_small_training_qualitative_ablation}
\vspace{-0.02in}

Below, we conduct a series of ablation studies to better understand how the proposed method works. Due to the space limit, we provide extended study in \cref{sec:app_exp_ablation}.

\vspace{0.02in}
\noindent\textbf{Bottleneck dimension.} In \cref{fig:ablation_gen_vs_bottleneck}, we show synthesized images from the models with varying bottleneck dimensions of the class-agnostic prompt generator. 
In \cref{fig:ablation_gen_vs_bottleneck_10img}, we use 10 target training images and the bottleneck dimension is chosen between 2$\sim$128. We see that our method behaves somewhat sensibly to the bottleneck dimension as discussed in \cref{sec:method_scdvpt}. When it is too small, it does not fully adapt to the target domain. On the other hand, when it is large, the class-agnostic prompt overpowers the class-specific prompt, and start to generate only target in-distribution images.

Similar behavior is observed in \cref{fig:ablation_gen_vs_bottleneck_1img} where we conduct an experiment using 1 target training image with varying dimension from 1$\sim$7. When the target images are less, it is recommended to use even fewer bottleneck units.

\vspace{0.02in}
\noindent\textbf{Attention control (\cref{eq:attn_ctrl}).} In \cref{fig:ablation_attn_ctrl_without}, we find that the class-agnostic prompt dominates without attention control and the model fails to synthesize images from novel classes, except for those close to the object category of the target domain training images (\eg, chihuahua, koala). On the other hand, the attention control promotes the usage of the class-specific prompt and transfer the domain knowledge effectively across novel classes as in \cref{fig:ablation_attn_ctrl_with}.

\vspace{0.02in}
\noindent\textbf{DA-CFG (\cref{sec:method_dacfg}).} We visualize {\expset} with varying guidance scales in \cref{fig:ablation_domadapt_cfg}. The model is trained on 8 images of ``Van Gogh house painting''. While we see the transfer of target domain knowledge already happening with no guidance, the object shapes are seriously distorted. With a larger guidance scale, the model synthesizes more articulated images of an object. 


%
Moreover, we confirm the importance of dropping class-agnostic prompt $p_{\phi}$ for domain adaptive synthesis. In \cref{fig:ablation_domadapt_cfg} second and fourth rows, we show images synthesized without dropping $p_{\phi}$ from the second term of \cref{eq:dacfg}. While the model generates more articulated images with larger guidance scales, it starts to generate images in the source domain, losing adaptation capability.

\begin{table*}[t]
    \centering
    \resizebox{0.9\linewidth}{!}{%
    \begin{tabular}{c|@{\hspace{0.1in}}c@{\hspace{0.1in}}|c|@{\hspace{0.1in}}c@{\hspace{0.1in}}|c|@{\hspace{0.1in}}c@{\hspace{0.1in}}}
        \toprule
        image & top 5 classes predicted by \textbf{A} &image & top 5 classes predicted by \textbf{A} & image & top 5 classes predicted by \textbf{A} \\
        \midrule
         \parbox[c]{0.5in}{\includegraphics[width=0.5in]{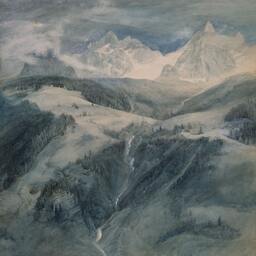}} & \makecell{Komodo dragon, Irish Wolfhound, \\cassette, \textbf{mountain}, \textbf{volcano}} & \parbox[c]{0.5in}{\includegraphics[width=0.5in]{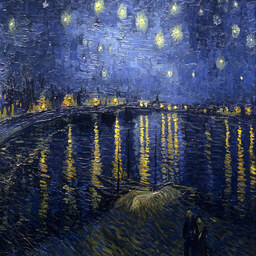}} & \makecell{\textbf{gondola}, king penguin, rapeseed, \\drilling rig, \textbf{paddle}} & \parbox[c]{0.5in}{\includegraphics[width=0.5in]{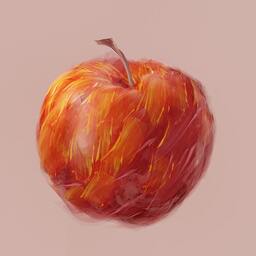}} & \makecell{plastic bag, \textbf{Granny Smith apple}, \\bee eater, \textbf{carved pumpkin}, lifeboat} \\
        \midrule
         \parbox[c]{0.5in}{\includegraphics[width=0.5in]{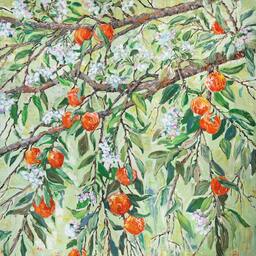}} & \makecell{\textbf{rose hip}, monarch butterfly, \\lorikeet, apron, canoe} & \parbox[c]{0.5in}{\includegraphics[width=0.5in]{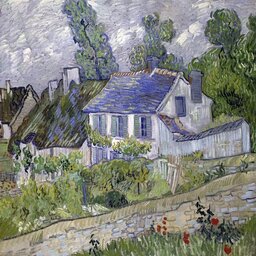}} & \makecell{\textbf{greenhouse}, maypole, \textbf{entertainment} \\\textbf{center}, Groenendael dog, \textbf{monastery}} & \parbox[c]{0.5in}{\includegraphics[width=0.5in]{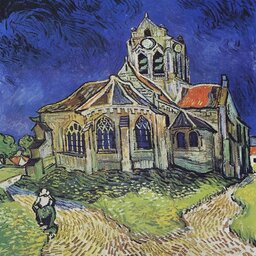}} & \makecell{\textbf{barn}, \textbf{entertainment center}, chameleon, \\steam locomotive, \textbf{castle}} \\
        \bottomrule
    \end{tabular}
    }
    \vspace{0.05in}
    \caption{Top 5 ImageNet classes by the class affinity matrix \textbf{A}. Relevant classes to the images are bold-faced.}
    \label{tab:class_prediction}
    \vspace{-0.1in}
\end{table*}

\begin{table*}[t]
    \centering
    \resizebox{0.9\linewidth}{!}{%
    \begin{tabular}{l|c|c|c|c|c|c|c|c|c|c|>{\columncolor[gray]{0.9}}c}
    \toprule
        Synthesis & art & cartoon & deviant art & embroidery & graffiti & graphic & origami & painting & sketch & toy & average \\
         \midrule
        Source & {92.0} & 134.4 & 89.0 & 166.3 & 166.5 & 138.8 & 155.5 & 78.4 & 94.0 & 97.7 & 121.3\\
        Source + StyTr$^{2}$~\cite{deng2022stytr2} & 70.8 & 99.6 & 70.7 & 158.9 & 114.2 & 111.5 & 141.8 & \textbf{52.6} & 84.6 & 82.2 & 98.7 \\
        \midrule
        Target, in-dist~\cite{sohn2022visual} & 168.0 & 118.3 & 173.8 & 176.1 & 168.0 & 182.0 & 142.9 & 145.1 & 163.1 & 133.3 & 157.1\\
        Target, zero-shot (single) & 95.5 & {85.9} & {78.2} & {121.3} & {114.0} & {109.3} & {116.7} & {64.0} & {47.5} & {84.2} & {91.6}\\
        Target, zero-shot (ensemble) & \textbf{67.0} & \textbf{67.4} & \textbf{68.8} & \textbf{115.7} & \textbf{89.0} & \textbf{100.7} & \textbf{106.1} & {65.1} & \textbf{36.8} & \textbf{74.9} & \textbf{79.1} \\
    \bottomrule
    \end{tabular}
    }
    \vspace{0.05in}
    \caption{FIDs (lower the better) on 10 target domains of the ImageNet-R~\cite{hendrycks2021many}. ``Source'' refers to the class-conditional synthesis of the original MaskGIT trained on the ImageNet, ``Target, in-dist'' refers to the synthesis conditioned on learned prompt and target domain in-distribution classes, and ``Target, zero-shot'' refers the proposed zero-shot synthesis conditioned on learned prompt and source classes. }
    \label{tab:imagenetr_fid}
    \vspace{-0.1in}
\end{table*}

\vspace{0.02in}
\noindent\textbf{Does accurate class prediction matter?} As our method is designed to disentangle the domain knowledge using the class information from the source domain, it seems plausible to assume that the target images should be recognized as one of classes of the source domain. We check the predicted classes of images in \cref{fig:zdais_oneshot_training} from the class affinity matrix \textbf{A}\footnote{This is done by ranking scores from $c^{\mbox{th}}$ row of \textbf{A}: $\mathbf{a}_{c}\,{=}\,[a_{j,c}]$.} and report top 5 prediction.
As in \cref{tab:class_prediction}, most images have relevant classes in their top 5 prediction (\eg, Van Gogh painting in the fourth row is predicted as gondola, whose typical images contain a boat in a river). This confirms that our method works as designed, but also suggests the limitation if the model fails to recognize object or concept in the image. We share such examples in \cref{sec:app_exp_ablation_prediction}.

\subsection{Quantitative Evaluation of ZDAIS}
\label{sec:exp_zdais_small_training_quantitative}

To test the efficacy of our method at scale, we need quantitative evaluation. Unfortunately, there is no benchmark to support our study yet. To this end, we propose a new benchmark using ImageNet-R~\cite{hendrycks2021many}. Below, we explain the dataset and settings, followed by results.

\vspace{0.02in}
\noindent\textbf{Dataset.} ImageNet-R~\cite{hendrycks2021many} is built to systematically study the robustness of the image classification models. As such, the dataset contains images from diverse domains, such as painting, sketch, deviant art, embroidery, or toy, while sharing categories with the ImageNet. Images are labeled with both the class and domain. We test on 10 domains.

\vspace{0.02in}
\noindent\textbf{Setting.} We pick 5 images from 5 classes for each domain and use them for visual prompt tuning. The list of images is in \cref{tab:supp_image_source_imagenetr}. We experiment with two settings. First, a single class-agnostic prompt is learned to model all 5 images (``single''). Second, we train one class-agnostic prompt per image and synthesize by randomly sampling prompt tokens (``ensemble''). Implementation details are in \cref{sec:app_exp_zdais_implementation details}.

\vspace{0.02in}
\noindent\textbf{Evaluation.} Images from each domain of the ImageNet-R are used as ground-truths to compute FIDs. Since the number of images in each domain varies from $550$ (origami) to $4634$ (sketch), FIDs are not comparable between domains. We generate 3 types of images -- source, target in-domain, and target zero-shot, for FID. Source refers to the synthesis by the pretrained MaskGIT, thus representing the distribution of an ImageNet~\cite{deng2009imagenet}. Target in-domain refers to the synthesis by conditioning on the target domain in-distribution classes used for visual prompt tuning, which is consistent with a typical setting for generative transfer learning~\cite{sohn2022visual}. Finally, target zero-shot refers to the {\expset}. 
%
Source classes are used as a condition only if there exist ground-truth images belonging to the same class in the target domain. We test 3 times and synthesize $20$k images for each run.

Furthermore, we evaluate the two step process of class-conditional image generation followed by NST, as introduced in \cref{sec:method_problem_formulation}, to show the difference between {\expset} and NST and how well our proposed method solves {\expset} problem. For NST, we employ state-of-the-art StyTr$^{2}$~\cite{deng2022stytr2}.

\begin{figure}[ht]
    \centering
    \begin{subfigure}[b]{0.49\linewidth}
        \centering
        \includegraphics[width=\textwidth]{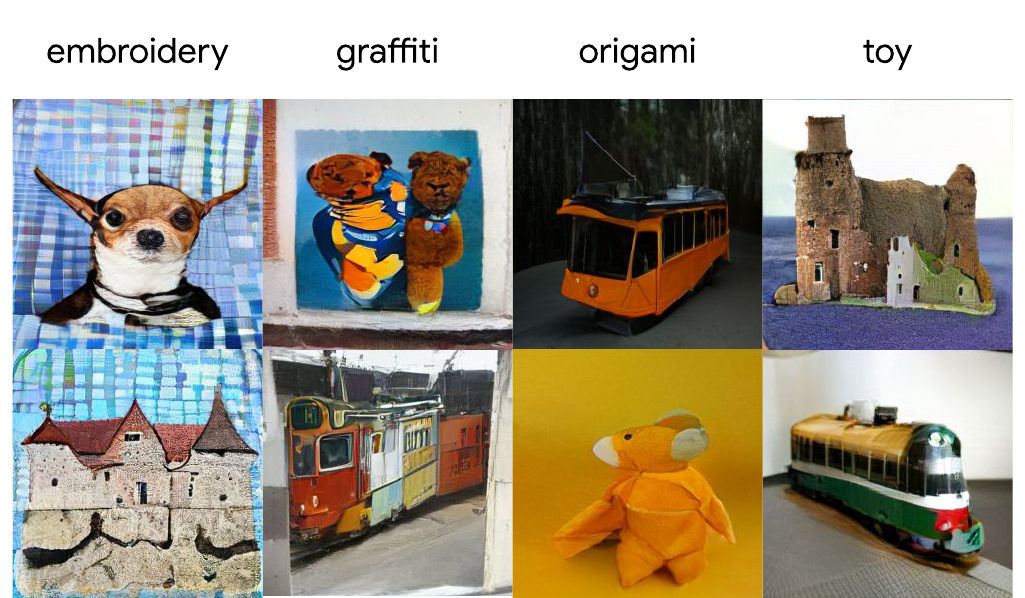}
        \caption{Ours}
        \label{fig:vis_imagenetr_ours}
    \end{subfigure}
    \begin{subfigure}[b]{0.49\linewidth}
        \centering
        \includegraphics[width=\textwidth]{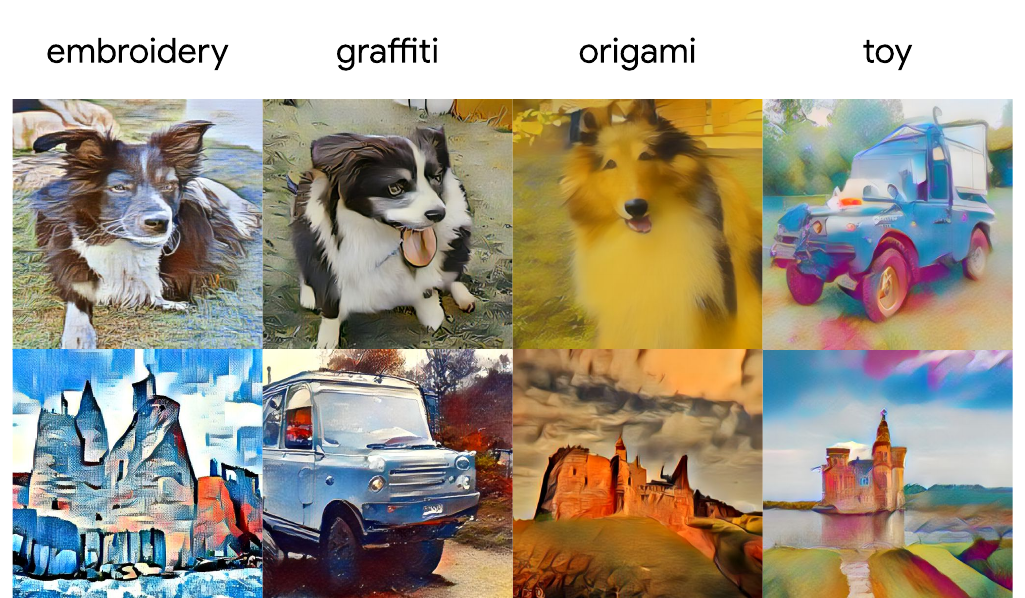}
        \caption{MaskGIT + StyTr$^{2}$~\cite{deng2022stytr2}.}
        \label{fig:vis_imagenetr_ours}
    \end{subfigure}
    \caption{Synthesis on selected target domains of ImageNet-R.}
    \vspace{-0.1in}
    \label{fig:vis_imagenetr}
\end{figure}

\vspace{0.02in}
\noindent\textbf{Results.} We report FIDs in \cref{tab:imagenetr_fid}.
%
%
We see that the proposed method (``Target, zero-shot'') achieves the lowest FID scores for most target domains. This confirms our previous findings in \cref{sec:exp_zdais_small_training_qualitative} that our method synthesizes not only good quality images, but also more faithful to each of the target domains than the MaskGIT without domain adaptive synthesis (``Source''). Note that this is not achievable with existing GTL methods~\cite{wang2020minegan,shahbazi2021efficient,sohn2022visual}. While NST (``Source + StyTr$^{2}$'') improves FIDs via style transfer to each of the target domains, they still fall short of our method.
Moreover, the ``ensemble'' approach results in improved FIDs, as it retains the generation diversity better, which is important for certain domains (\eg, art, cartoon, graffiti). Qualitative difference between two methods are shown in \cref{fig:vis_imagenetr}.
More images by our method are in \cref{fig:imagenetr_synthesis}.

%

\begin{table*}[t]
    \centering
    \resizebox{0.9\linewidth}{!}{%
    \begin{tabular}{c|c|c|c|c|c|c|c|c|c|c|c|c|>{\columncolor[gray]{0.9}}c}
    \toprule
        Source domain & \multicolumn{3}{c|}{Art} & \multicolumn{3}{c|}{Clipart} & \multicolumn{3}{c|}{Product} & \multicolumn{3}{c|}{Real} &  \\  
        Target domain & Clipart & Product & Real & Art & Product & Real & Art & Clipart & Real & Art & Clipart & Product & \multirow{-2}{*}{average} \\
        \midrule
        Source only & 60.9 & 75.1 & 86.5 & 66.8 & 70.9 & 76.5 & 64.2 & 57.9 & 83.7 & \textbf{81.8} & 66.7 & \textbf{88.9} & 73.3 \\
        Source + Synth (w/o adapt) & 61.8 & 84.3 & \textbf{87.8} & 77.4 & \textbf{83.2} & \textbf{87.2} & 77.4 & 62.8 & \textbf{89.3} & 80.7 & 63.4 & \textbf{88.9} & 78.7 \\
        Source + Synth (Ours) & \textbf{69.0} & \textbf{84.8} & \textbf{88.2} & \textbf{80.9} & \textbf{84.2} & \textbf{86.9} & \textbf{80.9} & \textbf{69.8} & \textbf{89.1} & \textbf{82.7} & 70.7 & \textbf{89.0} & \textbf{81.4} \\ \midrule
        \cite{jhoo2021collaborative} & \textbf{71.0} & 76.5 & 85.1 & 62.1 & 68.7 & 75.1 & 64.4 & \textbf{69.2} & 82.0 & 77.9 & \textbf{76.2} & \textbf{88.5} & 74.7 \\
    \bottomrule
    \end{tabular}
    }
    \vspace{0.05in}
    \caption{Zero-shot domain adaptation accuracy on Office-home dataset~\cite{venkateswara2017deep}. Numbers for \cite{jhoo2021collaborative} are taken from their paper. The best and those within standard error are bold-faced.}
    \label{tab:officehome}
    \vspace{-0.1in}
\end{table*}

\subsection{Zero-shot Domain Adaptation by Synthesis}
\label{sec:exp_zeroshot_adaptation}

We test the quality of \expset{} on zero-shot domain adaptation (ZSDA)~\cite{peng2018zero,jhoo2021collaborative}, a challenging task as there is no data available in the target domain for the task of interest to train a classifier. We approach this by using synthesized images from our method conditioned on the classes for the task of interest as a training data.
We conduct experiments on Office-home dataset~\cite{venkateswara2017deep} following the protocol in \cite{jhoo2021collaborative}, which we further detail in \cref{sec:app_exp_zsda}. 

Similarly to previous sections, we use a class-conditional MaskGIT trained on ImageNet as a source model and apply our method using images from each split of target domains for zero-shot image synthesis. Following \cite{jhoo2021collaborative}, we fine-tune ResNet-50~\cite{he2016deep} trained on the ImageNet on the combined source and synthesized images.

Results are shown in \cref{tab:officehome}. Compared to those trained on the source dataset only, classifiers trained on the combined source and synthesized images show consistent and significant improvement. In particular, we see 4$\sim$12\% gain in accuracy when the target domain is ``Clipart'', the most different from other domains, implying our method synthesizes images belonging to the target domain and classes. Visualization in \cref{fig:officehome_synthesis} further confirms our claim.
Moreover, our simple approach is shown to outperform state-of-the-art ZSDA method~\cite{jhoo2021collaborative} by a large margin.

\vspace{-0.05in}
\section{Related Work}
\label{sec:related}
\vspace{-0.02in}
To our knowledge, there is no previous work tackling the exact \expset{} setting. We review related works of similar capabilities and highlight the differences.

\vspace{0.02in}
\noindent\textbf{Generative Transfer Learning} aims to build a generative model for target tasks, which often have limited data for training, by transferring knowledge from the source model trained on a large dataset. Prior works~\cite{wang2020minegan,shahbazi2021efficient,ojha2021few,yang2021one} have been focusing on transferring from generative adversarial network (GAN)~\cite{goodfellow2014generative,brock2018large} and recently extended to other generative models~\cite{sohn2022visual}.
While they are shown to generate images belonging to the target domain, these methods are limited to synthesizing images of in-distribution whose training data is given. On the contrary, we study the composition of the semantic (\eg, object categories) from the source model and the style depicted by a few out-of-distribution images from unseen classes in the target domain.

Closest to {\expset}, cross-domain adaptation is studied in \cite{yang2021one} that transfers the style of a single target domain image of an irrelevant class. Their demonstration has been limited to synthesizing images of a certain class (\eg, church) while we show generation from diverse classes in the source class-conditional image generation model. We also make a direct comparison in \cref{fig:supp_compare_with_genda}, showing that our method learns to transfer higher-level concepts than color, tone or texture.

%

\huiwen{I feel we need to mention https://stylegan-nada.github.io/ and spliceViT somewhere. I would put it in this section, but then the claim may no longer hold. what do you think? }\kihyuk{Added stylegan-nada at the end of next paragraph. could you take another look? for spliceViT may fit better in neural style transfer paragraph?}

\vspace{0.02in}
\noindent\textbf{Zero-shot Image Synthesis with Text Guidance}~\cite{ramesh2021zero,ramesh2022hierarchical,nichol2021glide,rombach2022high,saharia2022photorealistic,yu2022scaling} is relevant to ours. The text-to-image models generate images by specifying the style in the text prompt, \eg, ``watercolor painting'', ``Van Gogh starry night''. They are shown effective for well known styles describable in text, but we point out two differences. 
First, they are trained from a large amount of training images of these styles.
For example, one can easily find more than hundred images of ``Van Gogh starry night'' or ``watercolor painting'' with variations from the LAION-400M dataset~\cite{schuhmann2021laion}.\footnote{We browse images using above keywords in \href{https://rom1504.github.io/clip-retrieval/}{clip-retrieval UI}.} In {\expset}, we show that learning composable token of such style is done from one image that is unseen in source model training. Second, there exist many modes in the style distribution (\eg, \cref{fig:watercolor_supp}), which may not be easily articulated with a text phrase. In practice, to capture such details with text-to-image models, one requires an extensive prompt engineering~\cite{oppenlaender2022taxonomy,liu2022design,hertz2022prompt} and careful tuning~\cite{kim2022diffusionclip,su2022dual,xia2021tedigan,patashnik2021styleclip,gal2022stylegan,crowson2022vqgan}. Nonetheless, as pointed out in \cite{gal2022stylegan}, text guidance may be inherently ambiguous, whereas such ambiguity could be resolved for our method as it is guided by reference images. 


%
Closest to ours include textual inversion~\cite{gal2022image} and model fine-tuning~\cite{ruiz2022dreambooth,kawar2022imagic} of text-to-image diffusion models~\cite{rombach2022high,saharia2022photorealistic}. \cite{gal2022image} proposes to learn a new vocabulary representing reference images by textual inversion. In DreamBooth~\cite{ruiz2022dreambooth}, a token is learned with a ground-truth class prior, showing a better decomposition of style and semantic information. While they leverage the compositional capability of text-to-image models, we propose novel techniques, such as attention control or domain adaptive CFG, to enhance disentanglement for class-conditional generative models.
Moreover, unlike \cite{ruiz2022dreambooth}, our method does not require model fine-tuning, making it not only efficient but also amenable to compose learned tokens. Our method is applicable to wide range of conditional generative models, \eg, text-conditioned generative transformers~\cite{yu2022scaling,villegas2022phenaki}.

%

\vspace{0.02in}
\noindent\textbf{Neural Style Transfer} (NST)~\cite{gatys2015neural,ghiasi2017exploring,deng2022stytr2,park2022styleformer,tumanyan2022splicing} translates a content image to a certain style characterized by the style reference image while maintaining the contents (\eg, structure, identity). As such, NST methods learn to disentangle content and style information with a set of losses for structure, identity preservation and appearance transfer, whose signals are given from discriminative models like VGG~\cite{simonyan2014very} or DINO~\cite{caron2021emerging}. Our method makes use of the semantic knowledge of pretrained generative transformers to disentangle target style, and is trained only with a masked token modeling loss. While both methods are used to synthesize images that transfers appearance or style, two methods have different goals and thus are not directly comparable. We provide a visual comparison in \cref{fig:app_comparision_to_nst} for interested readers.


\lu{The differences to the related works are not very obvious to me after reading the section. Perhaps adding a table summarizing their difference is helpful.}
\huiwen{Maybe you can directly point out that `we observe that our adapted outputs are usually capable of preserving more structural or semantic correspondence than neural style transfer methods'.
(just saying that the setup is different is a bit weak, talking about why the set up in the paper is more useful or powerful would be stronger here.  Additionally, If there is visual comparison in the experiment section with style transfer methods, it would be much stronger. )}

\vspace{-0.2in}
\section{Conclusion}
\label{sec:conclusion}
\vspace{-0.05in}
Towards better understanding of compositional generalization of models for image synthesis, we study the task of zero-shot domain adaptive image synthesis ({\expset}) in the space of a class-conditional MaskGIT~\cite{chang2022maskgit}. Through an extensive empirical study, we show that compositional generalization is achievable even without large amounts of training images nor rich text supervision. 
Moreover, we demonstrate that synthesized images by state-of-the-art generative models helps the classification on the challenging zero-shot domain adaptation setting.
The proposed techniques, such as source class distilled visual prompt tuning or domain adaptive classifier-free guidance, are generic, and we expect them to be readily applicable to other family of conditional generative models, such as text-to-image~\cite{yu2022scaling,villegas2022phenaki}.

{\small
\bibliographystyle{ieee_fullname}
\bibliography{reference}
}

\newpage
\onecolumn

\appendix

\begin{figure}[t]
\centering
\begin{lstlisting}[language=Python]
import flax.linen as nn
import jax.numpy as jnp

class ClassAgnosticTokenGenerator(nn.Module):
  d_embed: int  # Bottleneck dimension (P)
  d_token: int  # Token dimension (D)

  @nn.compact
  def __call__(self, ids: jnp.ndarray):
    """Calls class agnostic token generator.

    Args:
      ids: B x 1, values are all zeros.
         
    Return:
      jnp.ndarray of dimension B x 1 x D.
    """
    MLP_c = nn.Embed(1, self.d_embed)
    MLP_t = nn.Dense(self.d_token)

    return MLP_t(nn.LayerNorm(MLP_c(ids)))
    

class ClassDistilledTokenGenerator(nn.Module):
  n_class: int  # Number of target class (C_{tgt})

  @nn.compact
  def __call__(self, cls_ids: jnp.ndarray, W: jnp.ndarray):
    """Calls class distilled token generator.

    Args:
      cls_ids: B x 1, class labels.
      W: C_src x D, a matrix for class embedding tokens.
         
    Return:
      jnp.ndarray of dimension B x 1 x D.
    """
    Affn = nn.Embed(self.n_class_tgt, W.shape[0])
    
    token = jax.nn.softmax(Affn(cls_ids)) * W  # B x D
    return token[:, None, :]


def compute_attention_weight(query, class_token, domain_token):
  """Computes attention weights with attention control.
  
  Args:
    query: tensor of size D x sequence length
    class_token: tensor of size D x 1
    domain_token: tensor of size D x 1
  
  Return:
    tensor of size (sequence length) x (sequence length + 2)
  """
  key = jnp.concatenate((class_token, domain_token, query), axis=1)
  attn_weight = jnp.einsum('dq,dk->qk', query, key)
  attn_weight = jnp.concatenate((jnp.maximum(attn_weight[:, 0], attn_weight[:, 1]), attn_weight[:, 1:], axis=1)
  return attn_weight

\end{lstlisting}
\caption{An example code for the token generator and the attention control of \cref{eq:attn_ctrl} in Flax-ish~\cite{flax2020github} format.}
\label{fig:prompt_generator}
\end{figure}

\begin{table}[t]
    \centering
    \resizebox{0.8\linewidth}{!}{%
    \begin{tabular}{c|l}
    \toprule
        Dataset & Link \\ \midrule
        \multirow{8}{*}{Van Gogh House painting}& \url{https://www.rawpixel.com/image/3868934/illustration-image-art-vincent-van-gogh-person} \\
& \url{https://www.rawpixel.com/image/3865273/illustration-image-art-vincent-van-gogh-house} \\
& \url{https://www.rawpixel.com/image/3866294/illustration-image-art-vincent-van-gogh-house} \\
& \url{https://www.rawpixel.com/image/3868302/illustration-image-art-vincent-van-gogh-house} \\
& \url{https://www.rawpixel.com/image/3864574/illustration-image-art-vincent-van-gogh} \\
& \url{https://www.rawpixel.com/image/3864611/illustration-image-art-vincent-van-gogh-house} \\
& \url{https://www.rawpixel.com/image/537424/free-illustration-image-van-gogh-factory} \\
& \url{https://www.rawpixel.com/image/537422/free-illustration-image-van-gogh-cottage} \\\midrule
\multirow{2}{*}{Van Gogh starry night}& \url{https://search-production.openverse.engineering/image/ec5f5215-1307-457f-bc14-3e17d4fa4735} \\
& \url{https://search-production.openverse.engineering/image/278dd07a-0b55-416b-b1f5-05798f64cf34} \\\midrule
\multirow{9}{*}{Haunted house}& \url{https://search-production.openverse.engineering/image/17662d45-aa50-474e-83b6-bd170eda9bd9} \\
& \url{https://search-production.openverse.engineering/image/f010aca4-14d0-4464-a706-b66ed7fb8569} \\
& \url{https://search-production.openverse.engineering/image/ce8bd8d7-a509-4ece-a9e2-eb8581d0fb00} \\
& \url{https://search-production.openverse.engineering/image/2f1a1eee-c033-4781-bad0-3e12212a2361} \\
& \url{https://search-production.openverse.engineering/image/89e86034-fe58-4f32-8259-9dbc4bac8ebc} \\
& \url{https://search-production.openverse.engineering/image/089c1cb0-f070-4d1f-9fc5-a82a7939d411} \\
& \url{https://www.rawpixel.com/image/5906049/photo-image-public-domain-house-halloween} \\
& \url{https://www.rawpixel.com/image/5964852/free-public-domain-cc0-photo} \\
& \url{https://www.rawpixel.com/image/6051791/free-public-domain-cc0-photo} \\\midrule
\multirow{10}{*}{Abandoned cars}& \url{https://search-production.openverse.engineering/image/9e826ffa-bac1-4892-b78a-04eded1cefcf} \\
& \url{https://search-production.openverse.engineering/image/545bb05c-cafc-42fc-89cc-34b0e51593a2} \\
& \url{https://search-production.openverse.engineering/image/b1e4bc4e-982e-4792-a513-bdea0a5f72cb} \\
& \url{https://search-production.openverse.engineering/image/b103ce3b-c74c-41c4-97e0-8c3a7563bd59} \\
& \url{https://www.rawpixel.com/image/3090749/free-photo-image-abandoned-car-vintage} \\
& \url{https://www.rawpixel.com/image/5941829/free-public-domain-cc0-photo} \\
& \url{https://www.rawpixel.com/image/4026101/oldsmobile-route-66} \\
& \url{https://search-production.openverse.engineering/image/fbb2922a-a03c-44e7-9571-6901bba27957} \\
& \url{https://search-production.openverse.engineering/image/f579052d-d3c8-4a59-8217-97b9f4b6ae55} \\
& \url{https://search-production.openverse.engineering/image/f8db9513-f0a6-40e0-9e23-e6e34b5e04f3} \\\midrule
\multirow{10}{*}{Watercolor dogs}& \url{https://pixy.org/6458158/} \\
& \url{https://pixy.org/5790662/} \\
& \url{https://pixy.org/6379346/} \\
& \url{https://pixy.org/5792209/} \\
& \url{https://pixy.org/5792070/} \\
& \url{https://pixy.org/5787524/} \\
& \url{https://pixy.org/5782945/} \\
& \url{https://pixy.org/6487827/} \\
& \url{https://pixy.org/6557843/} \\
& \url{https://pixy.org/5788893/} \\\midrule
\multirow{6}{*}{chinese inkpainting}& \url{https://search-production.openverse.engineering/image/382e0452-14a5-4895-aff2-917dfec8f40c} \\
& \url{https://search-production.openverse.engineering/image/22a3f474-128d-4a69-8d4e-98f904a77170} \\
& \url{https://search-production.openverse.engineering/image/3aa5eee8-3a49-4e8e-9df9-f6ca2e0dc01c} \\
& \url{https://search-production.openverse.engineering/image/0793e761-3437-4591-a0e9-5a8f4d53ea7f} \\
& \url{https://search-production.openverse.engineering/image/293879fc-9f71-410f-b96b-161493e14377} \\
& \url{https://search-production.openverse.engineering/image/04edb78b-b848-430b-9166-c61efab08a3c} \\\midrule
\multirow{10}{*}{aurora mountain}& \url{https://unsplash.com/photos/CgoRzWX4CDg} \\
& \url{https://unsplash.com/photos/U_diPCXCBxU} \\
& \url{https://unsplash.com/photos/-OkHUsepnzw} \\
& \url{https://unsplash.com/photos/pDeagUyN-Pk} \\
& \url{https://unsplash.com/photos/ZJDMls6ppY8} \\
& \url{https://unsplash.com/photos/Hn8N4I4eHA0} \\
& \url{https://unsplash.com/photos/fpaSXDuoHkc} \\
& \url{https://unsplash.com/photos/58X3XfxxevU} \\
& \url{https://unsplash.com/photos/uWmWoH9maR4} \\
& \url{https://unsplash.com/photos/w1yDuFs-kGY} \\\midrule
\multirow{3}{*}{inkpainting mountain}& \url{https://unsplash.com/photos/6fv0MEf3FUE} \\
& \url{https://unsplash.com/photos/NspHfyZnMBE} \\
& \url{https://unsplash.com/photos/Vc8GBqapdfs} \\\midrule
\multirow{1}{*}{oilpainting apple}& \url{https://unsplash.com/photos/LQTdG9SJpyA} \\\midrule
\multirow{1}{*}{oilpainting appletree}& \url{https://unsplash.com/photos/K3QvdUkcQp4} \\\midrule
\multirow{4}{*}{sketch mushroom}& \url{https://freesvg.org/mushrooms-vector-illustration} \\
& \url{https://freesvg.org/psm-v07-d144-common-meadow-mushroom} \\
& \url{https://freesvg.org/1517757743} \\
& \url{https://freesvg.org/puffball} \\\midrule
\multirow{13}{*}{watercolor images}& \url{https://unsplash.com/photos/KRztl5I6xac} \\
& \url{https://unsplash.com/photos/0pJPixfGfVo} \\
& \url{https://unsplash.com/photos/YIfFVwDcgu8} \\
& \url{https://unsplash.com/photos/9dnNnTrHxmI} \\
& \url{https://unsplash.com/photos/Tyg0rVhOTrE} \\
& \url{https://unsplash.com/photos/8D-0K6JUAEE} \\
& \url{https://unsplash.com/photos/6NSVToSYwV0} \\
& \url{https://unsplash.com/photos/-KfLa4I4eTo} \\
& \url{https://unsplash.com/photos/-IAS_N85adA} \\
& \url{https://unsplash.com/photos/X2QwsspYk_0} \\
& \url{https://unsplash.com/photos/TAZga9MibgA} \\
& \url{https://unsplash.com/photos/wvD0zZnRbcw} \\
& \url{https://unsplash.com/photos/6dY9cFY-qTo} \\
    \bottomrule
    \end{tabular}
    }
    \caption{Image sources for experiments in \cref{sec:exp_zdais_small_training_qualitative}.}
    \label{tab:supp_image_source}
\end{table}
\iffalse
\begin{table*}[t]
    \centering
    \resizebox{0.4\linewidth}{!}{%
    \begin{tabular}{c|l}
    \toprule
    Domain & File name \\ \midrule
         \multirow{5}{*}{art} & \texttt{n01443537/art\_10.jpg}\\
        &\texttt{n01855672/art\_0.jpg}\\
        &\texttt{n02129604/art\_4.jpg}\\
        &\texttt{n02814860/art\_15.jpg}\\
        &\texttt{n03452741/art\_0.jpg}\\\midrule
        \multirow{5}{*}{cartoon} & \texttt{n01443537/cartoon\_28.jpg}\\
        &\texttt{n01855672/cartoon\_3.jpg}\\
        &\texttt{n02129604/cartoon\_4.jpg}\\
        &\texttt{n02814860/cartoon\_22.jpg}\\
        &\texttt{n03452741/cartoon\_2.jpg}\\\midrule
        \multirow{5}{*}{deviantart} & \texttt{n01443537/deviantart\_16.jpg}\\
        &\texttt{n01855672/deviantart\_11.jpg}\\
        &\texttt{n02129604/deviantart\_23.jpg}\\
        &\texttt{n02814860/deviantart\_20.jpg}\\
        &\texttt{n03452741/deviantart\_0.jpg}\\\midrule
        \multirow{5}{*}{embroidery} & \texttt{n01443537/embroidery\_1.jpg}\\
        &\texttt{n01855672/embroidery\_0.jpg}\\
        &\texttt{n02129604/embroidery\_7.jpg}\\
        &\texttt{n02814860/embroidery\_5.jpg}\\
        &\texttt{n07614500/embroidery\_10.jpg}\\\midrule
        \multirow{5}{*}{graffiti} & \texttt{n01443537/graffiti\_1.jpg}\\
        &\texttt{n01855672/graffiti\_0.jpg}\\
        &\texttt{n02129604/graffiti\_9.jpg}\\
        &\texttt{n02814860/graffiti\_0.jpg}\\
        &\texttt{n07614500/graffiti\_1.jpg}\\\midrule
        \multirow{5}{*}{graphic} & \texttt{n01443537/graphic\_3.jpg}\\
        &\texttt{n01855672/graphic\_1.jpg}\\
        &\texttt{n02129604/graphic\_1.jpg}\\
        &\texttt{n02980441/graphic\_6.jpg}\\
        &\texttt{n03452741/graphic\_0.jpg}\\\midrule
        \multirow{5}{*}{origami} & \texttt{n01443537/origami\_9.jpg}\\
        &\texttt{n01855672/origami\_1.jpg}\\
        &\texttt{n02129604/origami\_1.jpg}\\
        &\texttt{n02814860/origami\_1.jpg}\\
        &\texttt{n03452741/origami\_3.jpg}\\\midrule
        \multirow{5}{*}{painting} & \texttt{n01443537/painting\_11.jpg}\\
        &\texttt{n01855672/painting\_0.jpg}\\
        &\texttt{n02129604/painting\_13.jpg}\\
        &\texttt{n02814860/painting\_14.jpg}\\
        &\texttt{n03452741/painting\_7.jpg}\\\midrule
        \multirow{5}{*}{sketch} & \texttt{n01443537/sketch\_14.jpg}\\
        &\texttt{n01855672/sketch\_10.jpg}\\
        &\texttt{n02129604/sketch\_0.jpg}\\
        &\texttt{n02814860/sketch\_1.jpg}\\
        &\texttt{n03452741/sketch\_21.jpg}\\\midrule
        \multirow{5}{*}{toy} & \texttt{n01443537/toy\_14.jpg}\\
        &\texttt{n01855672/toy\_4.jpg}\\
        &\texttt{n02129604/toy\_4.jpg}\\
        &\texttt{n02814860/toy\_1.jpg}\\
        &\texttt{n03452741/toy\_7.jpg}\\
    \bottomrule
    \end{tabular}
    }
    \caption{Image sources for ImageNet-R experiments in \cref{sec:exp_zdais_small_training_quantitative}.}
    \label{tab:supp_image_source_imagenetr}
\end{table*}
\else
\begin{table*}[t]
    \centering
    \resizebox{0.95\linewidth}{!}{%
    \begin{tabular}{@{\hspace{0.1in}}c@{\hspace{0.1in}}|l@{\hspace{0.75in}}||@{\hspace{0.1in}}c@{\hspace{0.1in}}|l@{\hspace{0.75in}}}
    \toprule
    Domain & File name & Domain & File name \\ \midrule
\multirow{5}{*}{art} & \texttt{n01443537/art\_10.jpg}&\multirow{5}{*}{graphic} & \texttt{n01443537/graphic\_3.jpg}\\
&\texttt{n01855672/art\_0.jpg}&&\texttt{n01855672/graphic\_1.jpg}\\
&\texttt{n02129604/art\_4.jpg}&&\texttt{n02129604/graphic\_1.jpg}\\
&\texttt{n02814860/art\_15.jpg}&&\texttt{n02980441/graphic\_6.jpg}\\
&\texttt{n03452741/art\_0.jpg}&&\texttt{n03452741/graphic\_0.jpg}\\\midrule
\multirow{5}{*}{cartoon} & \texttt{n01443537/cartoon\_28.jpg}&\multirow{5}{*}{origami} & \texttt{n01443537/origami\_9.jpg}\\
&\texttt{n01855672/cartoon\_3.jpg}&&\texttt{n01855672/origami\_1.jpg}\\
&\texttt{n02129604/cartoon\_4.jpg}&&\texttt{n02129604/origami\_1.jpg}\\
&\texttt{n02814860/cartoon\_22.jpg}&&\texttt{n02814860/origami\_1.jpg}\\
&\texttt{n03452741/cartoon\_2.jpg}&&\texttt{n03452741/origami\_3.jpg}\\\midrule
\multirow{5}{*}{deviantart} & \texttt{n01443537/deviantart\_16.jpg}&\multirow{5}{*}{painting} & \texttt{n01443537/painting\_11.jpg}\\
&\texttt{n01855672/deviantart\_11.jpg}&&\texttt{n01855672/painting\_0.jpg}\\
&\texttt{n02129604/deviantart\_23.jpg}&&\texttt{n02129604/painting\_13.jpg}\\
&\texttt{n02814860/deviantart\_20.jpg}&&\texttt{n02814860/painting\_14.jpg}\\
&\texttt{n03452741/deviantart\_0.jpg}&&\texttt{n03452741/painting\_7.jpg}\\\midrule
\multirow{5}{*}{embroidery} & \texttt{n01443537/embroidery\_1.jpg}&\multirow{5}{*}{sketch} & \texttt{n01443537/sketch\_14.jpg}\\
&\texttt{n01855672/embroidery\_0.jpg}&&\texttt{n01855672/sketch\_10.jpg}\\
&\texttt{n02129604/embroidery\_7.jpg}&&\texttt{n02129604/sketch\_0.jpg}\\
&\texttt{n02814860/embroidery\_5.jpg}&&\texttt{n02814860/sketch\_1.jpg}\\
&\texttt{n07614500/embroidery\_10.jpg}&&\texttt{n03452741/sketch\_21.jpg}\\\midrule
\multirow{5}{*}{graffiti} & \texttt{n01443537/graffiti\_1.jpg}&\multirow{5}{*}{toy} & \texttt{n01443537/toy\_14.jpg}\\
&\texttt{n01855672/graffiti\_0.jpg}&&\texttt{n01855672/toy\_4.jpg}\\
&\texttt{n02129604/graffiti\_9.jpg}&&\texttt{n02129604/toy\_4.jpg}\\
&\texttt{n02814860/graffiti\_0.jpg}&&\texttt{n02814860/toy\_1.jpg}\\
&\texttt{n07614500/graffiti\_1.jpg}&&\texttt{n03452741/toy\_7.jpg}\\\bottomrule
\end{tabular}
    }
    \vspace{0.05in}
    \caption{Image sources for ImageNet-R experiments in \cref{sec:exp_zdais_small_training_quantitative}. For reader's convenience, \texttt{n01443537} is for ``goldfish'', \texttt{n01855672} for ``goose'', \texttt{n02129604} for ``tiger'', \texttt{n02814860} for ``lighthouse'', \texttt{n03452741} for ``grand piano'', \texttt{n02980441} for ``castle'', and \texttt{n07614500} for ``ice cream''.}
    \label{tab:supp_image_source_imagenetr}
\end{table*}
\fi

\begin{figure*}[t]
    \centering
    \includegraphics[width=\textwidth]{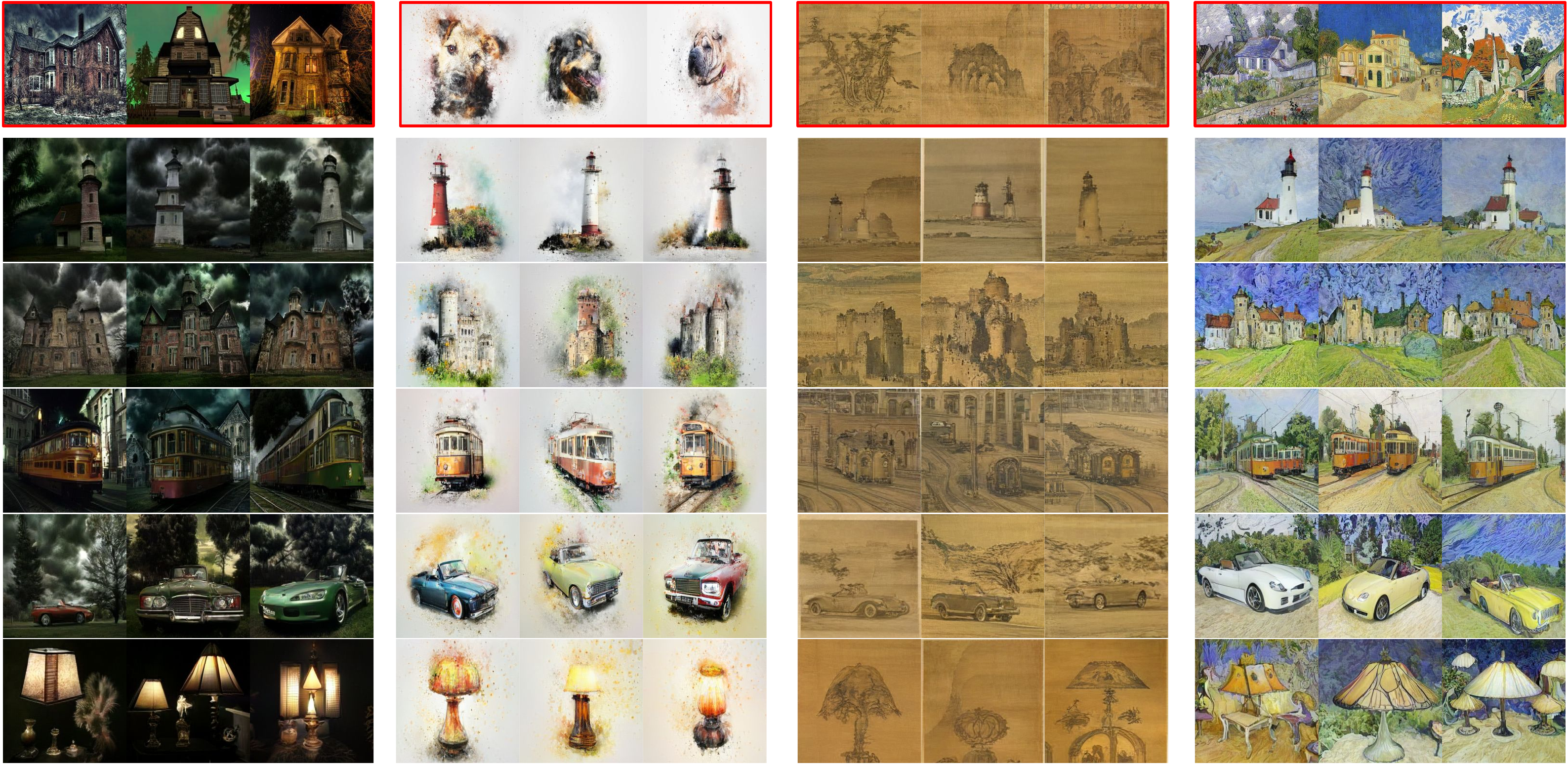}\\
    \vspace{0.1in}
    \includegraphics[width=\textwidth]{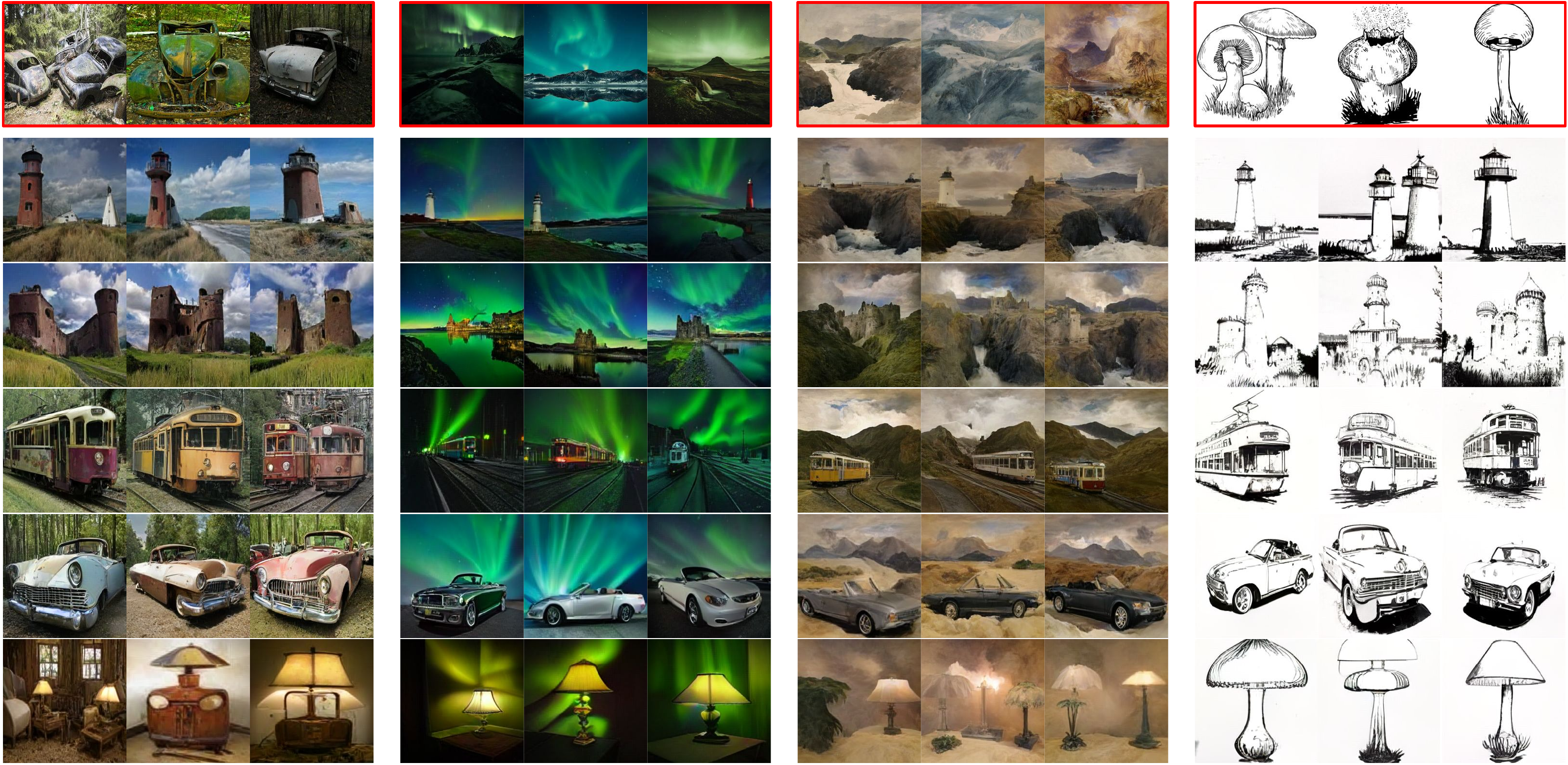}
    \caption{Zero-shot domain adaptive image synthesis from a few training images. Images inside red boxes are 3 training images (out of 6$\sim$10 in total) for each target domain, and the rest are synthesized. The class condition for each row is ``light house'', ``castle'', ``tram'', ``convertible'', and ``table lamp''. We provide information on training images in \cref{sec:app_image_source}.}
    \label{fig:zdais_fewshot_training_supp}
\end{figure*}

\begin{figure*}[t]
    \centering
    \includegraphics[width=\textwidth]{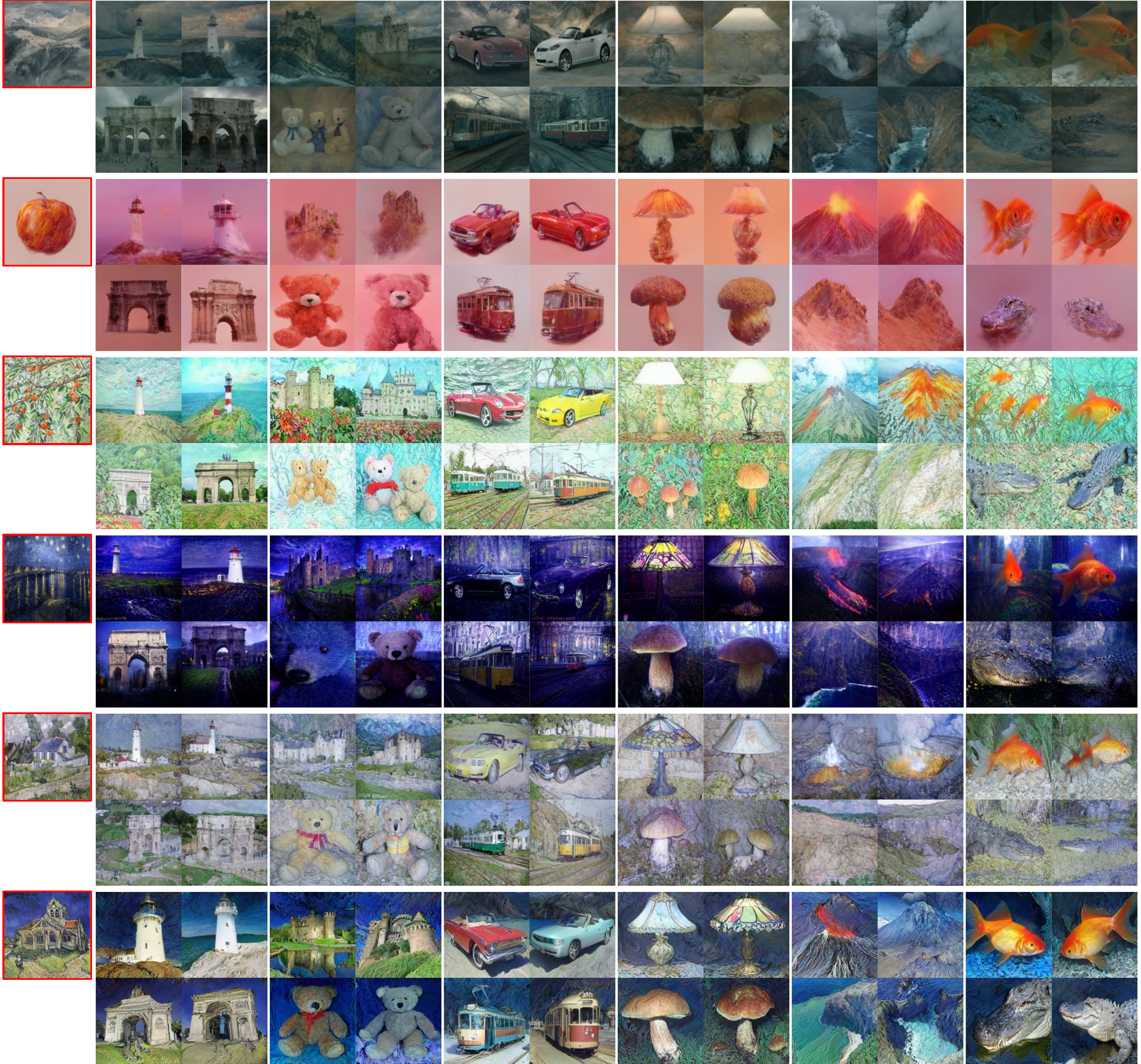}
    \caption{Zero-shot domain adaptive image synthesis from a single training image. Images inside red boxes are a \textbf{single (and only)} training image for each target domain, and the rest are synthesized. From left to right, top to bottom, class conditions are ``lighthouse'', ``castle'', ``convertible'', ``table lamp'', ``volcano'', ``goldfish'', ``triumphal arch'', ``teddy bear'', ``tram'', ``bolete'', ``cliff'', ``American alligator''.}
    \label{fig:app_zdais_oneshot_training_more}
\end{figure*}

\begin{figure*}[t]
\centering
    \includegraphics[width=\textwidth]{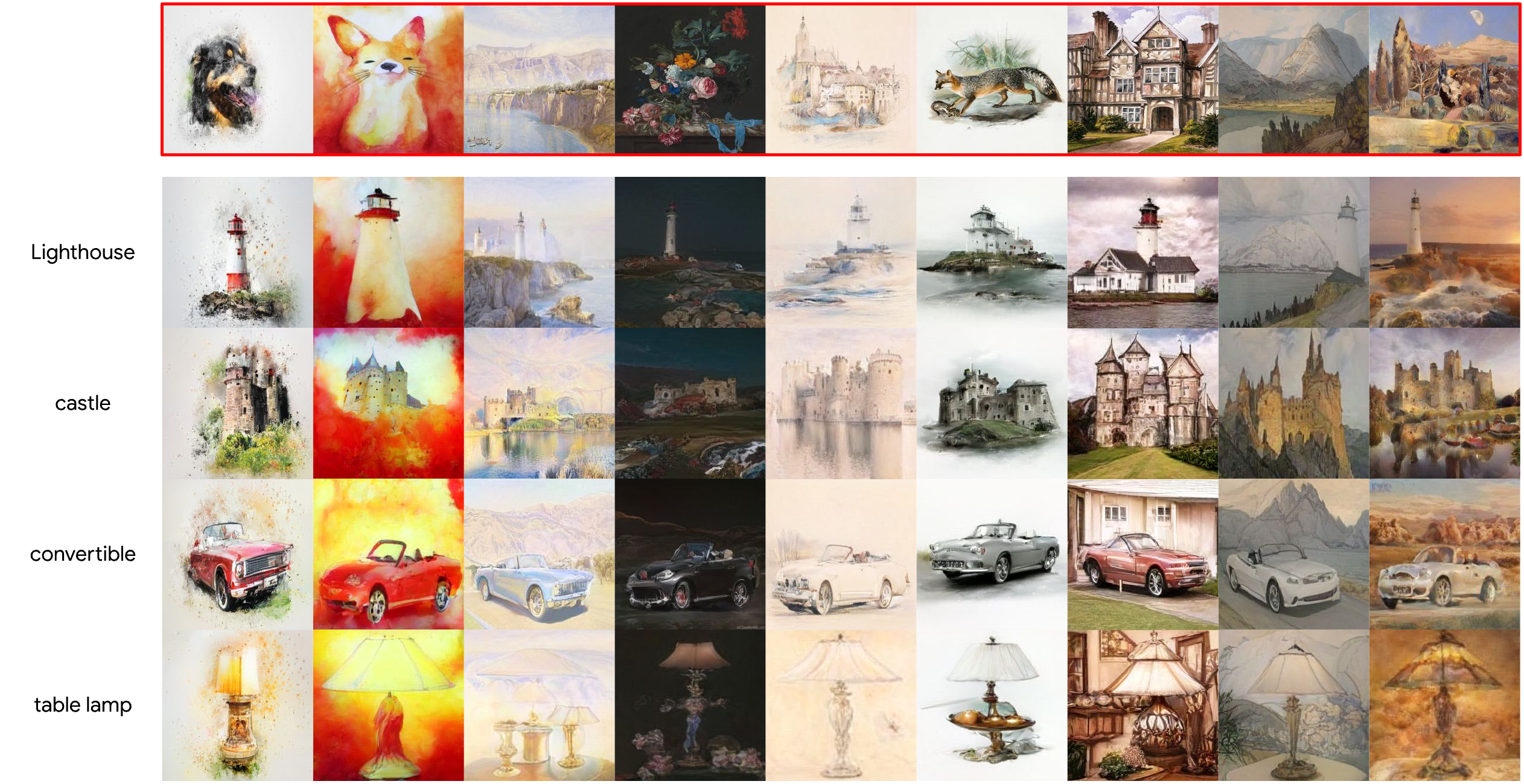}
    \caption{Many styles of ``watercolor painting''. Our method is able to adapt to different styles of ``watercolor painting'' using a single reference image per target domain (in red boxes) and generalize to unseen concepts. This is in contrast with text-guided image synthesis methods where inherent ambiguity could occur due to ambiguity in natural language prompt.}
    \label{fig:watercolor_supp}
\end{figure*}

\begin{figure*}[t]
    \centering
    \includegraphics[width=0.99\textwidth]{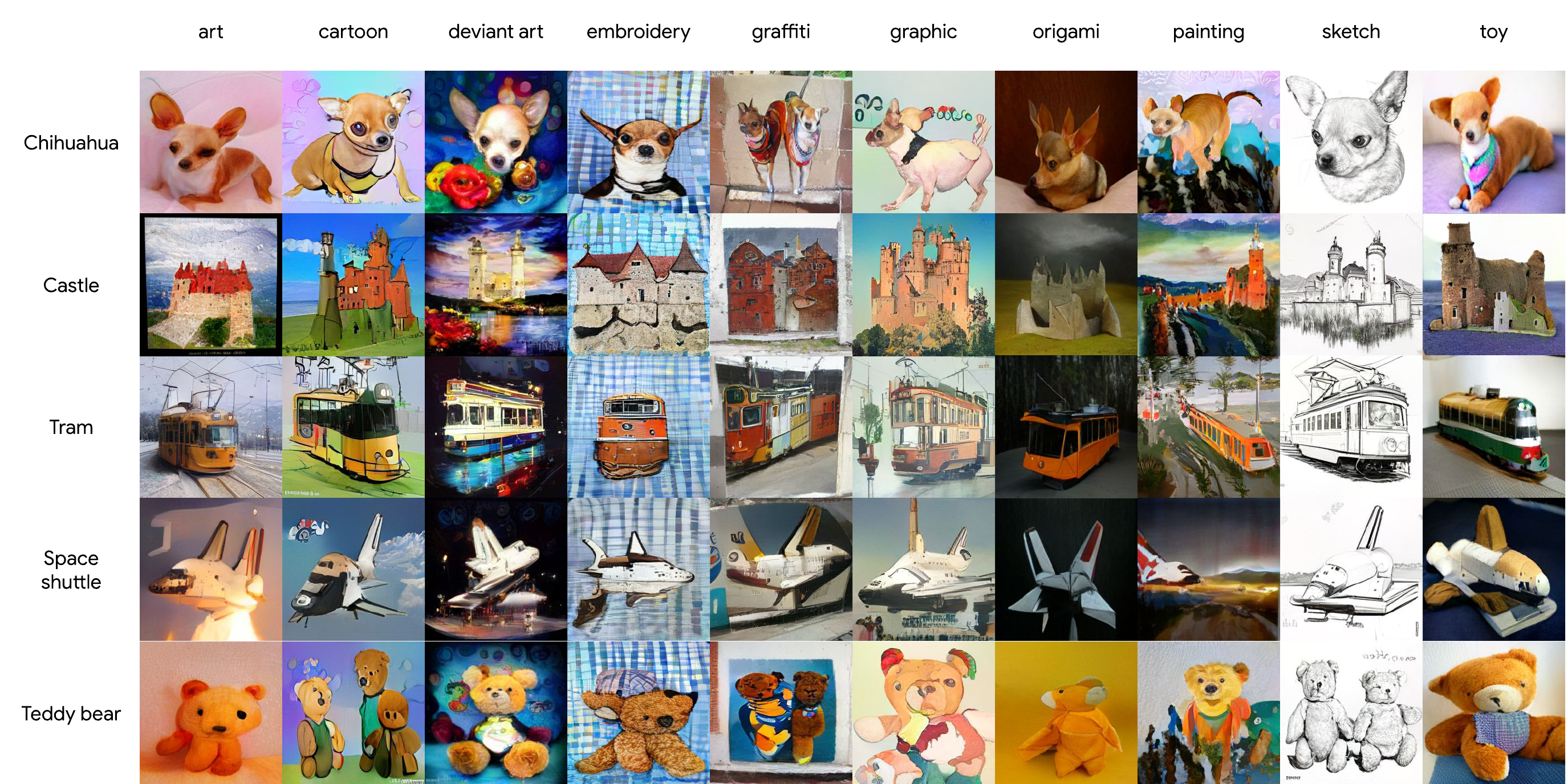}
    \caption{Zero-shot domain adaptive image synthesis on ImageNet-R~\cite{hendrycks2021many}.}
    \label{fig:imagenetr_synthesis}
\end{figure*}
\begin{figure*}[t]
    \centering
    \begin{subfigure}[b]{0.24\textwidth}
        \centering
        \includegraphics[width=\textwidth]{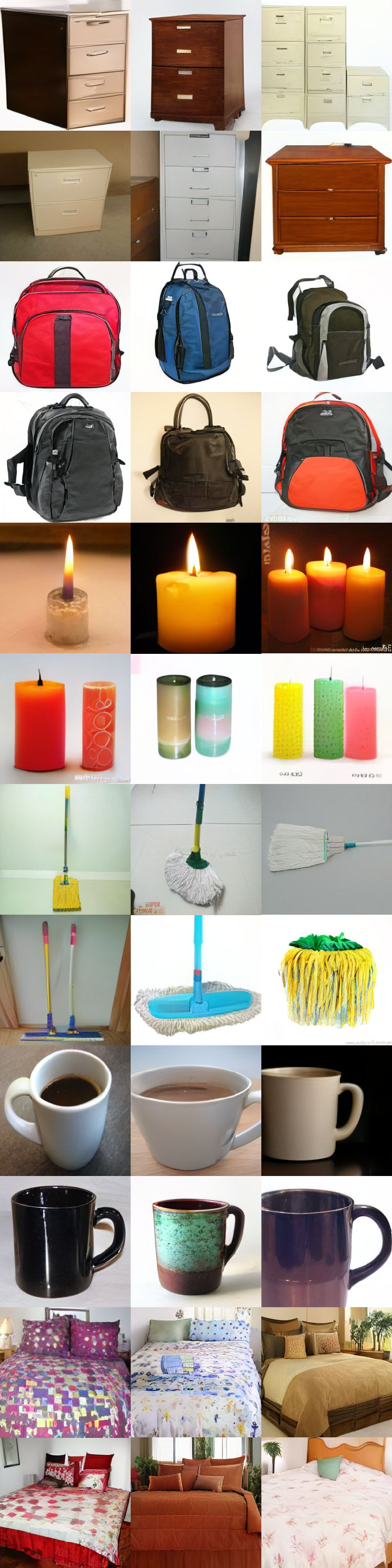}
        \caption{\textsc{Real World}}
        \label{fig:officehome_synthesis_real}
    \end{subfigure}
    \begin{subfigure}[b]{0.24\textwidth}
        \centering
        \includegraphics[width=\textwidth]{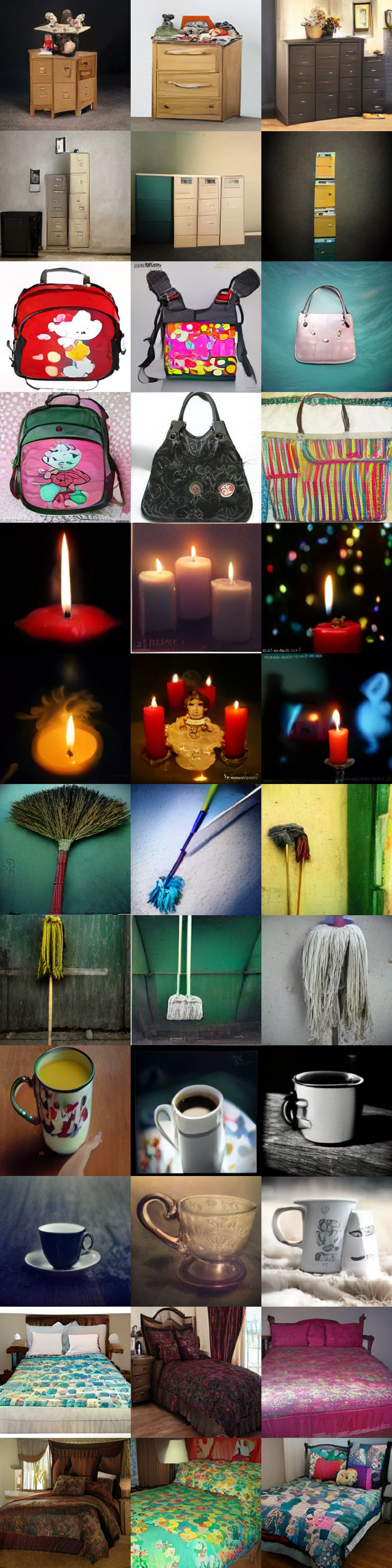}
        \caption{\textsc{Art}}
        \label{fig:officehome_synthesis_art}
    \end{subfigure}
    \begin{subfigure}[b]{0.24\textwidth}
        \centering
        \includegraphics[width=\textwidth]{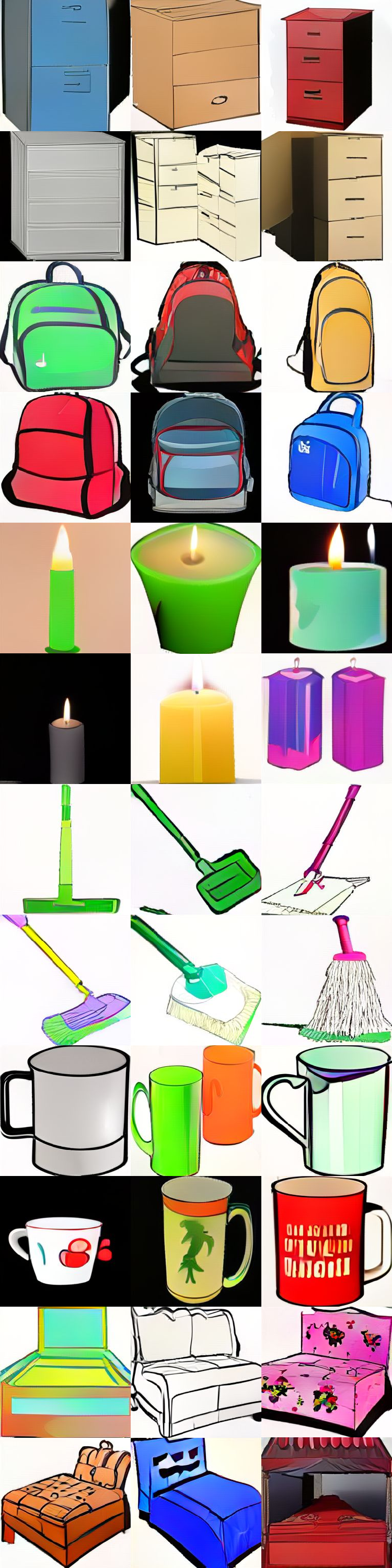}
        \caption{\textsc{Clipart}}
        \label{fig:officehome_synthesis_clipart}
    \end{subfigure}
    \begin{subfigure}[b]{0.24\textwidth}
        \centering
        \includegraphics[width=\textwidth]{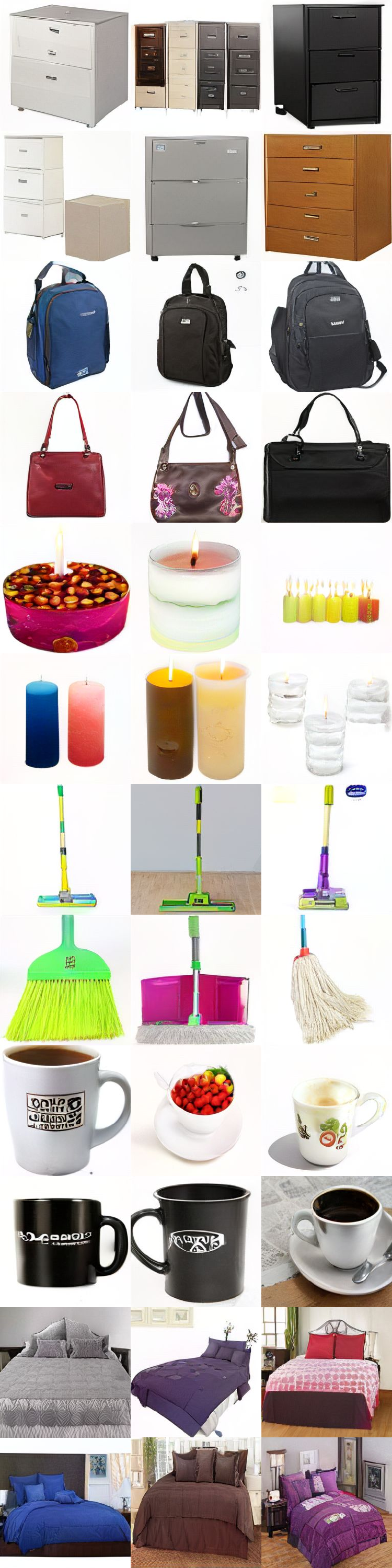}
        \caption{\textsc{Product}}
        \label{fig:officehome_synthesis_product}
    \end{subfigure}
    \caption{Zero-shot domain adaptive image synthesis on Office-home~\cite{venkateswara2017deep}. From top to bottom, class condition for each two rows are ``File cabinet'', ``Backpack'', ``Candle'', ``Mop'', ``Mug'', ``Bed''.}
    \label{fig:officehome_synthesis}
\end{figure*}

\begin{figure*}[ht]
    \centering
    \begin{subfigure}[b]{0.49\textwidth}
        \centering
        \includegraphics[width=0.99\textwidth]{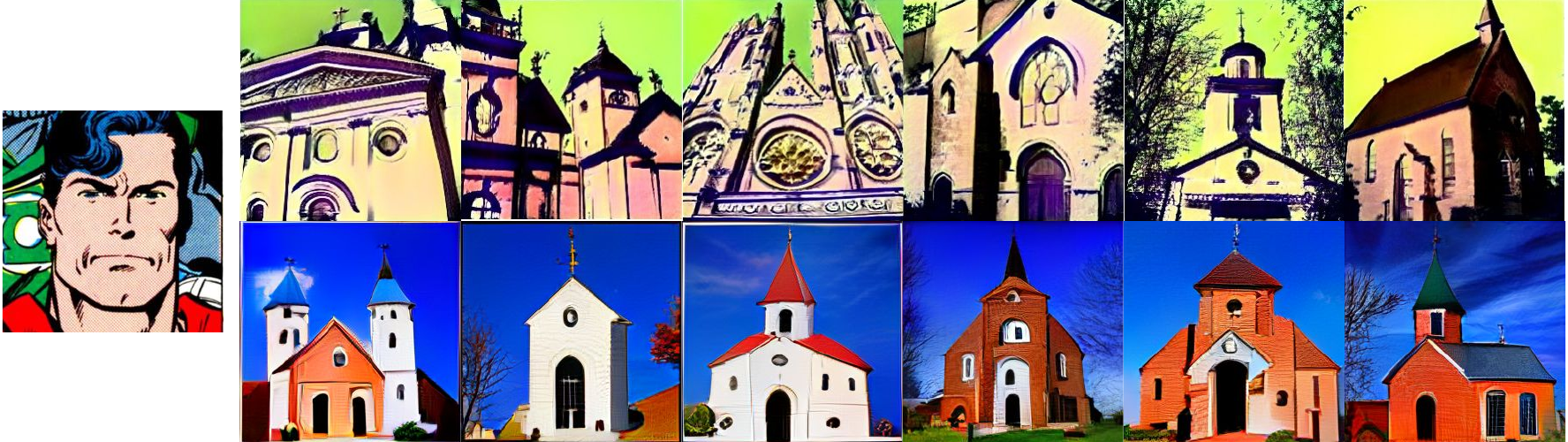}
        \caption{Superman}
        \label{fig:supp_compare_with_genda_superman}
    \end{subfigure}
    \begin{subfigure}[b]{0.49\textwidth}
        \centering
        \includegraphics[width=0.99\textwidth]{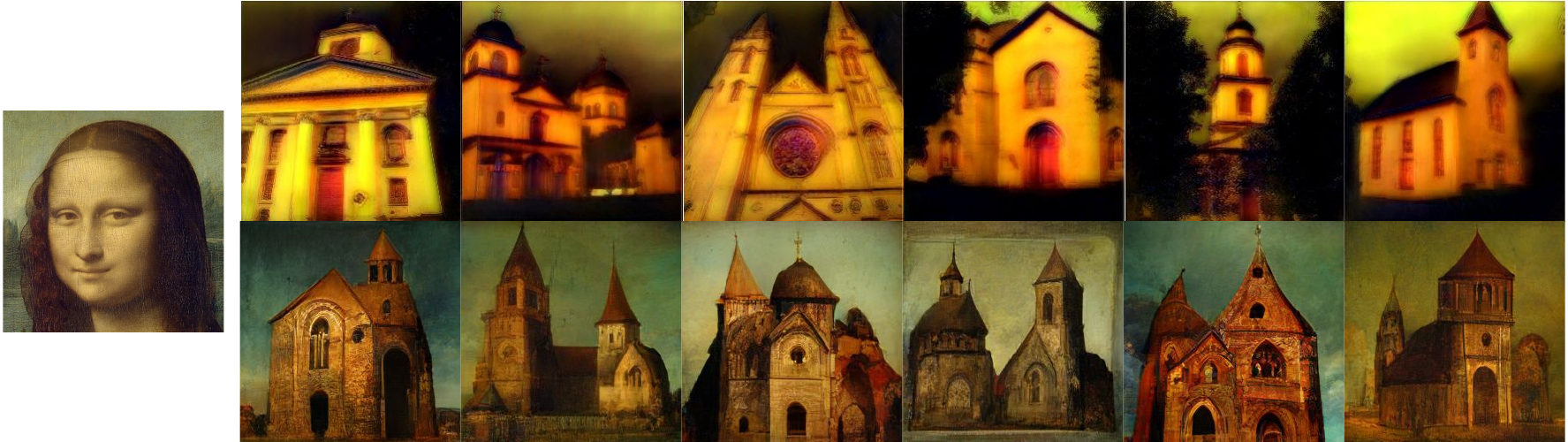}
        \caption{Mona Lisa}
        \label{fig:supp_compare_with_genda_mona_lisa}
    \end{subfigure}
    \caption{Qualitative comparison with GenDA~\cite{yang2021one}. Images in the first row of each figure are taken from \cite{yang2021one}, and images in the second row of each figure are synthesized by our method. Images on the left (superman, Mona Lisa) are given as target domain images.}
    \label{fig:supp_compare_with_genda}
\end{figure*}

\begin{figure}[t]
    \centering
    \includegraphics[width=\linewidth]{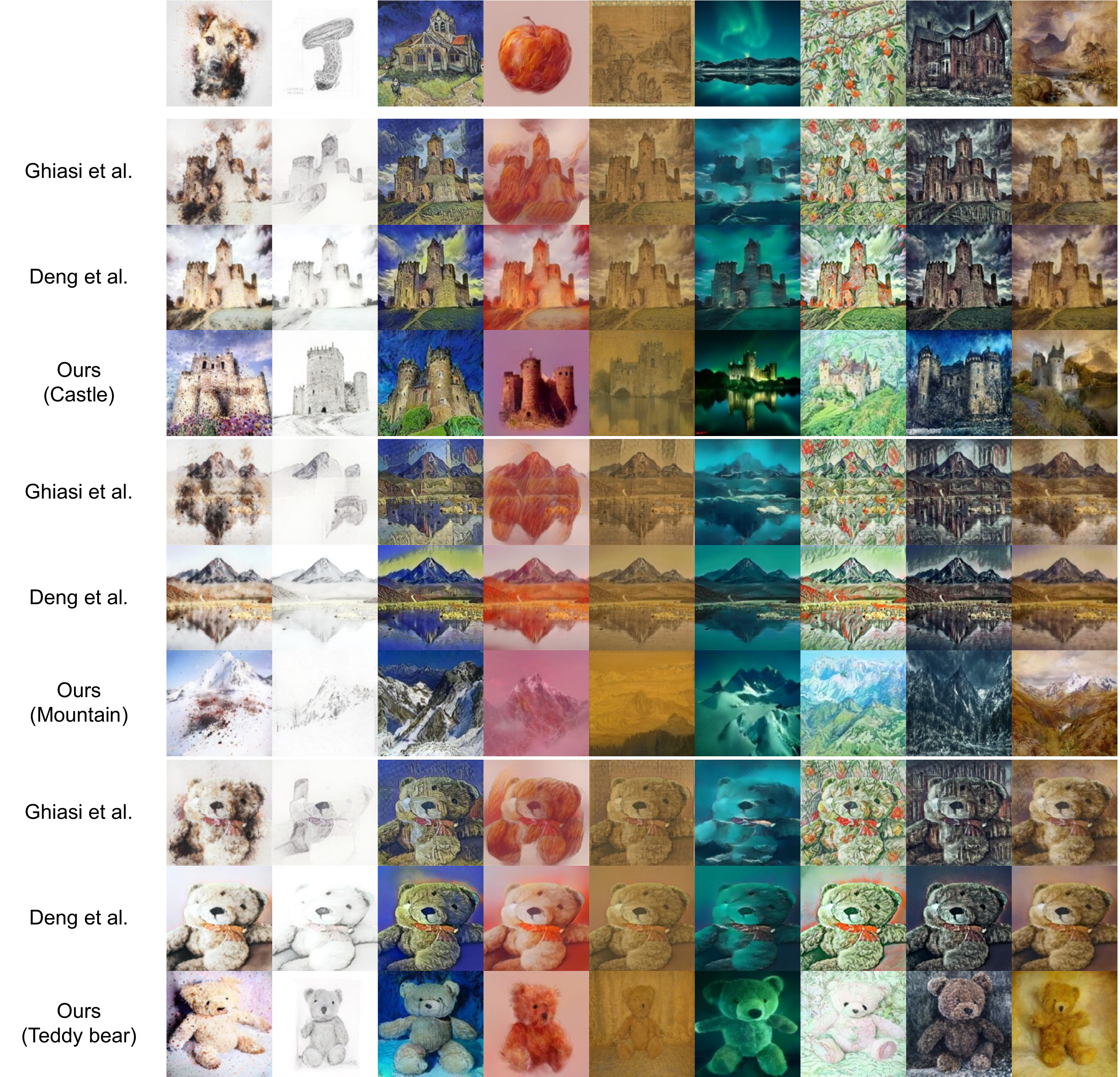}
    \caption{Comparison to neural style transfer methods of Ghiasi \etal.~\cite{ghiasi2017exploring} and Deng \etal.~\cite{deng2022stytr2}. Unlike neural style transfer methods whose input to the system consists of content and style images, our method does not take a content image as an input to enforce structural or identity-wise similarity between content and synthesized images. Instead, our method is class-conditional, \ie, input to the system is composed of class label and style image. Conditioned classes used here are ``castle'', ``mountain'', and ``teddy bear''.}
    \label{fig:app_comparision_to_nst}
\end{figure}

\clearpage

\section{Extended Descriptions on Method}
\label{sec:app_method}

\begin{figure}[h]
    \centering
    \includegraphics[width=0.7\textwidth]{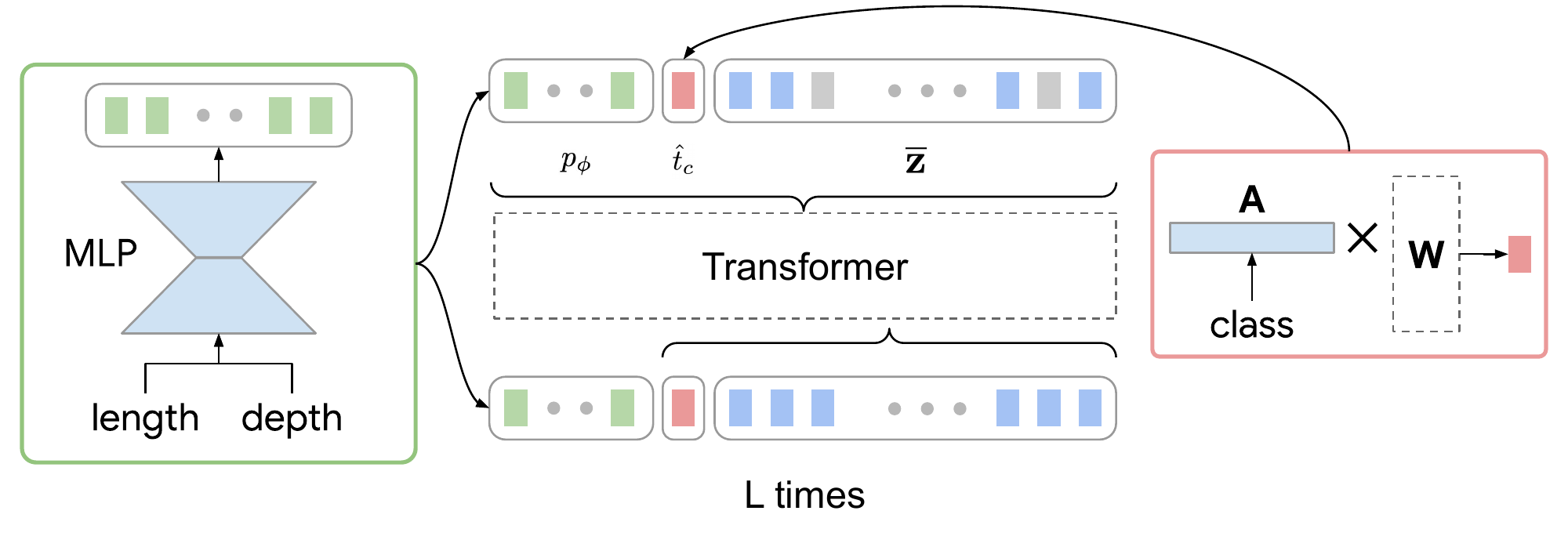}
    \caption{Generative visual prompt tuning with deep prompts. Separate learnable prompts (green) are provided for each transformer layer.}
    \label{fig:app_method_overview_detail}
\end{figure}

Here, we provide more detailed information on our proposed method. Especially, similarly to \cite{jia2022visual}, we find that the deep prompt, where separate learnable prompts are in place for all transformer layers, is more adaptive to the target domain than a shallow prompt. This is in contrast with \cite{sohn2022visual} where only the shallow prompt has been discussed. \cref{fig:app_method_overview_detail} illustrates deep prompt used in our method. Systematic study comparing different prompt designs for generative transfer is beyond the scope of this work.

\section{Experiments}
\label{sec:app_exp}

\subsection{Implementation Detail}
\label{sec:app_exp_zdais_implementation details}

\vspace{0.02in}
\noindent\textbf{Experiments in \cref{sec:exp_zdais_small_training_qualitative,sec:exp_zdais_small_training_quantitative}.} All experiments are done using the class-conditional MaskGIT~\cite{chang2022maskgit} trained on the ImageNet dataset, whose checkpoint is publicly available.\footnote{\url{https://github.com/google-research/maskgit}} We use a deep prompt, as explained in \cref{sec:app_method}, with $S\,{=}\,1$. For optimization, we use an Adam optimizer~\cite{kingma2014adam} with the learning rate of $0.001$ decaying with a cosine learning rate decay~\cite{loshchilov2016sgdr}. No weight decay nor learning rate warm-up is used. The parameters are updated for $1000$ steps with the batch size of $128$. We note that the number of training images are 1$\sim$10 for experiments of \cref{sec:exp_zdais_small_training_qualitative}, and as such we repeat the training data by applying a stochastic data augmentation including random crop and horizontal flip. However, using a large batch size is not essential, and we empirically find that using batch size of $16$ with $1000$ optimization steps is sufficient for our method to work. 
The code will be released publicly for reproducible research.

\subsection{{\expset} with Many Training Images}
\label{sec:app_exp_zdais_many_training}

Our prompt design in \cref{sec:method_scdvpt} is geared towards solving the problem when there are only a few training images provided from the target domain. As shown in \cref{sec:exp_zdais_small_training_qualitative_ablation}, we need a stronger regularization, \eg, reducing the bottleneck dimension, as there are fewer training images available. On the other hand, we hypothesize that the failure of disentanglement could be mitigated when there are many training images from diverse classes in the target domain.

\vspace{0.02in}
\noindent\textbf{Setting.} We conduct experiments to confirm our hypothesis, where we use an entire data of ImageNet-R sketch~\cite{hendrycks2021many} as a target domain training set, and test the \expset{} performance on the ImageNet-sketch~\cite{wang2019learning}. Note that ImageNet-sketch and ImageNet-R sketch are collected from two different groups and images in these two datasets do not necessarily overlap. Since ImageNet-R sketch contains $200$ overlapping classes to the ImageNet~\cite{deng2009imagenet}, we exclude images from those classes when computing the FID. The total number of images used for visual prompt tuning is $4,634$, \eg, $\approx$23 images per class. In addition, we conduct experiments using $1\,{\sim}\,5$ images per class ($200\,{\sim}\,1,000$ images in total, respectively) to see the impact of number of images per class on generation performance.

We train our models with varying class-agnostic token length $S\,{=}\,\{1,16,64,128\}$ without the bottleneck layer. To accommodate class-agnostic prompt of length longer than 1, we extend an attention control mechanism in \cref{eq:attn_ctrl} as follows:
\begin{equation}
    \mathrm{softmax}\big({\footnotesize\texttt{cat}}\boldsymbol{(}\max\big\{\mathbf{q}^{\top}\hat{t}_{c},\log\sum_{s=1}^{S}\exp\left(\mathbf{q}^{\top}p_{\phi,s}\right)\big\},\mathbf{q}^{\top}p_{\phi},\mathbf{q}^{\top}\mathbf{q}\boldsymbol{)}\big)\label{eq:attn_ctrl_ext}
\end{equation}
where we replace $\mathbf{q}^{\top}p_{\phi}$ into $\log\sum_{s=1}^{S}\exp\left(\mathbf{q}^{\top}p_{\phi,s}\right)$ from \cref{eq:attn_ctrl}. This ensures sufficient attention to be given to the class-specific token even when the class-agnostic prompt has multiple tokens.

\begin{figure*}[t]
    \centering
    \begin{subfigure}[b]{0.24\textwidth}
        \centering
        \includegraphics[width=\textwidth]{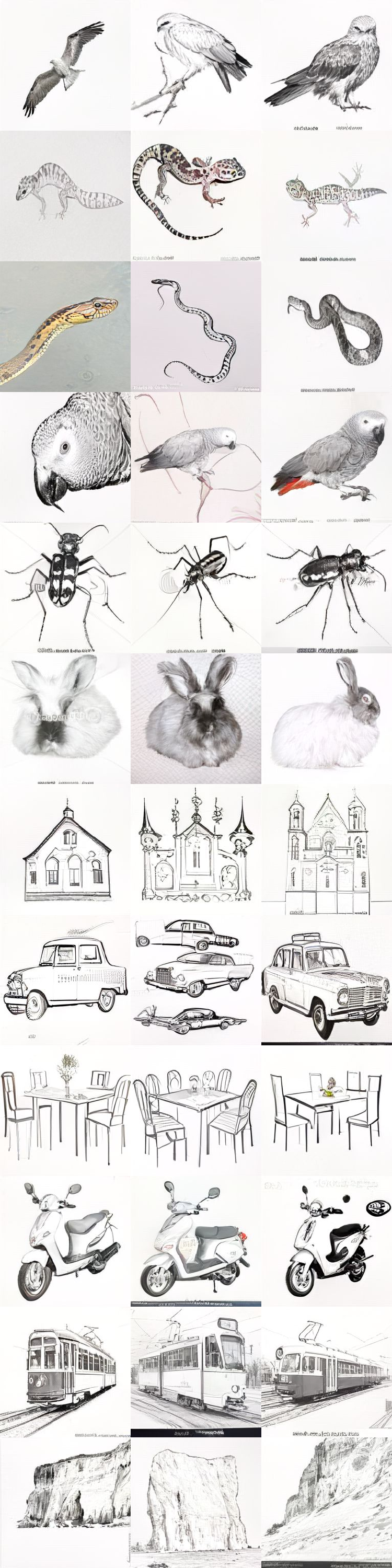}
        \caption{$S\,{=}\,1$}
        \label{fig:app_imagenetr_to_sketch_synthesis_s1}
    \end{subfigure}
    \hspace{0.02in}
    \begin{subfigure}[b]{0.24\textwidth}
        \centering
        \includegraphics[width=\textwidth]{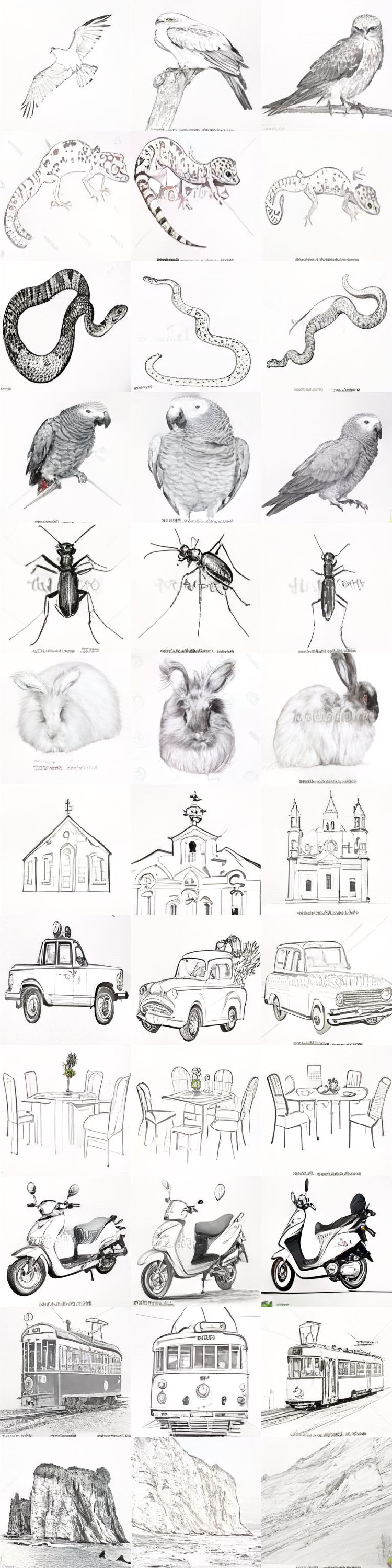}
        \caption{$S\,{=}\,16$}
        \label{fig:app_imagenetr_to_sketch_synthesis_s16}
    \end{subfigure}
    \hspace{0.02in}
    \begin{subfigure}[b]{0.24\textwidth}
        \centering
        \includegraphics[width=\textwidth]{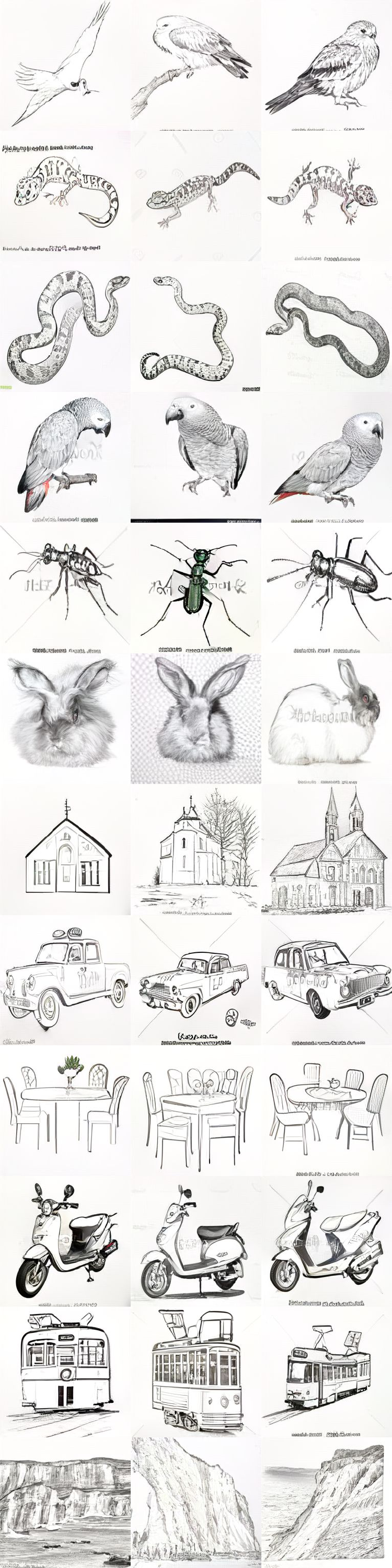}
        \caption{$S\,{=}\,64$}
        \label{fig:app_imagenetr_to_sketch_synthesis_s64}
    \end{subfigure}
    \hspace{0.02in}
    \begin{subfigure}[b]{0.24\textwidth}
        \centering
        \includegraphics[width=\textwidth]{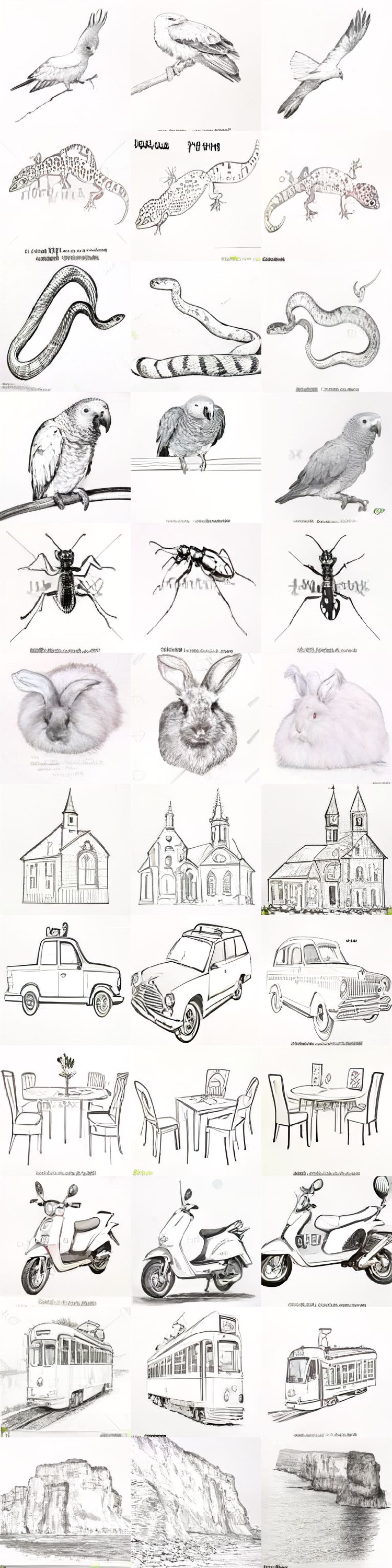}
        \caption{$S\,{=}\,128$}
        \label{fig:app_imagenetr_to_sketch_synthesis_s128}
    \end{subfigure}
    \caption{Zero-shot domain adaptive image synthesis trained on ImageNet-R sketch~\cite{hendrycks2021many}. Class conditions are ``kite (\texttt{n01608432})'', ``banded gecko (\texttt{n01675722})'', ``water snake (\texttt{n01737021})'', ``African grey parrot (\texttt{n01817953})'', ``tiger beetle (\texttt{n02165105})'', ``Angora rabbit (\texttt{n02328150})'', ``church (\texttt{n03028079})'', ``cab (\texttt{n02930766})'', ``dining table (\texttt{n03201208})'', ``motor scooter (\texttt{n03791053})'', ``streetcar (\texttt{n04335435})'', and ``cliff (\texttt{n09246464})'', which do not exist in the ImageNet-R sketch.}
    \label{fig:app_imagenetr_to_sketch_synthesis}
\end{figure*}

\vspace{0.02in}
\noindent\textbf{Result.} We report results in \cref{tab:imagenetr_to_sketch}. FIDs are computed between $20$k synthesized images and $20$k images sampled from classes that do not belong to ImageNet-R sketch dataset. Similarly to \cref{tab:imagenetr_fid}, ``Source'' refers to the synthesis of zero-shot classes using pretrained class-conditional MaskGIT without domain transfer, ``Target, in-dist'' refers to the in-distribution synthesis, \ie, domain adaptive synthesis conditioned on the classes of ImageNet-R sketch, and ``Target, zero-shot'' refers to the zero-shot domain adaptive image synthesis.
The FID measured by synthetic images of the source model is $77.0$, implying that there exists a significant domain gap between the ImageNet and ImageNet sketch. When measured between target in-distribution synthetic images and the target zero-shot ground-truth images, we still get significantly lower FIDs around $30$. This denotes that the FID is more affected by the domain than the class. Our proposed method for \expset{} further reduces the FID, achieving $14.8$ using 12 synthesis steps and $12.2$ using 36 synthesis steps. This suggests that our method synthesizes images not only belonging to the target distribution, but also respecting the class semantics. We visualize synthesized images from the zero-shot classes (\ie, which do not belong to classes of ImageNet-R sketch) in \cref{fig:app_imagenetr_to_sketch_synthesis}.

Different from experiments in \cref{sec:exp_zdais_small_training_qualitative} where the number of training images from the target domain is less than $10$, we observe a successful disentanglement of domain and class without strongly regularizing the prompt capacity. On the contrary, we see an improvement in the domain adaptive zero-shot synthesis with longer class-agnostic prompts. This implies that when there are many training examples, it is still beneficial to have a powerful class-agnostic prompt to adapt to the target domain.

\begin{figure*}[t]
    \centering
    \begin{subfigure}[b]{\textwidth}
        \centering
        \includegraphics[width=\textwidth]{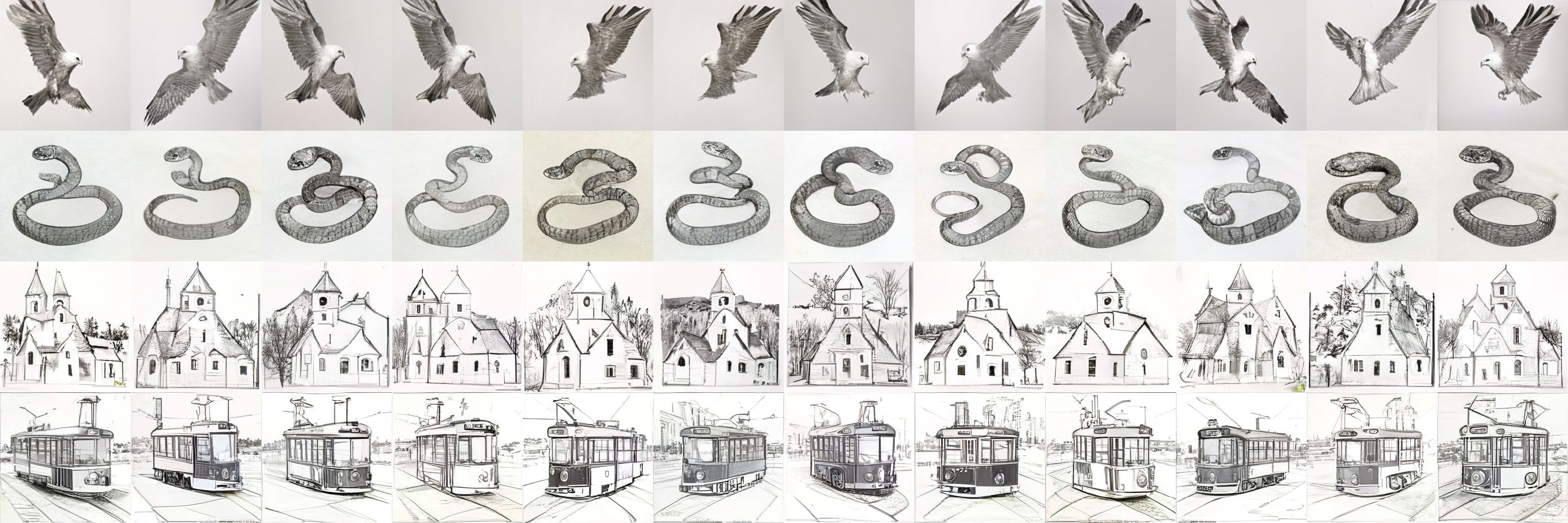}
        \caption{1 image per class.}
        \label{fig:app_imagenetr_to_sketch_synthesis_more_1img}
    \end{subfigure}
    \begin{subfigure}[b]{\textwidth}
        \centering
        \includegraphics[width=\textwidth]{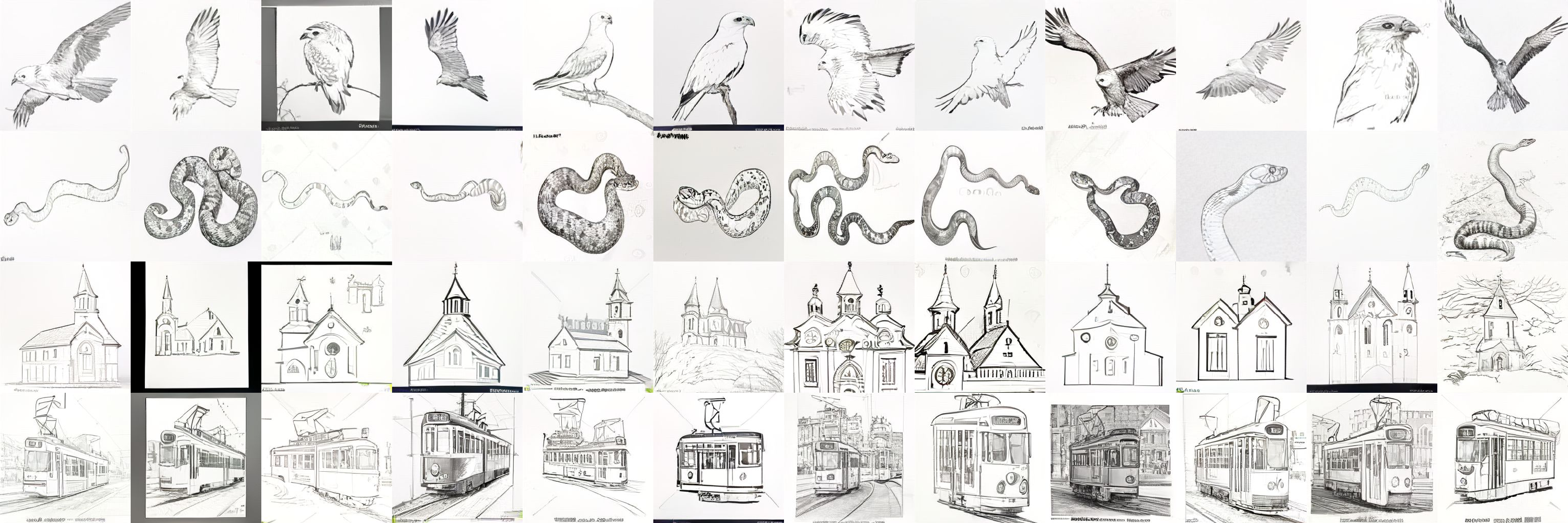}
        \caption{5 images per class.}
        \label{fig:app_imagenetr_to_sketch_synthesis_more_5img}
    \end{subfigure}
    \begin{subfigure}[b]{\textwidth}
        \centering
        \includegraphics[width=\textwidth]{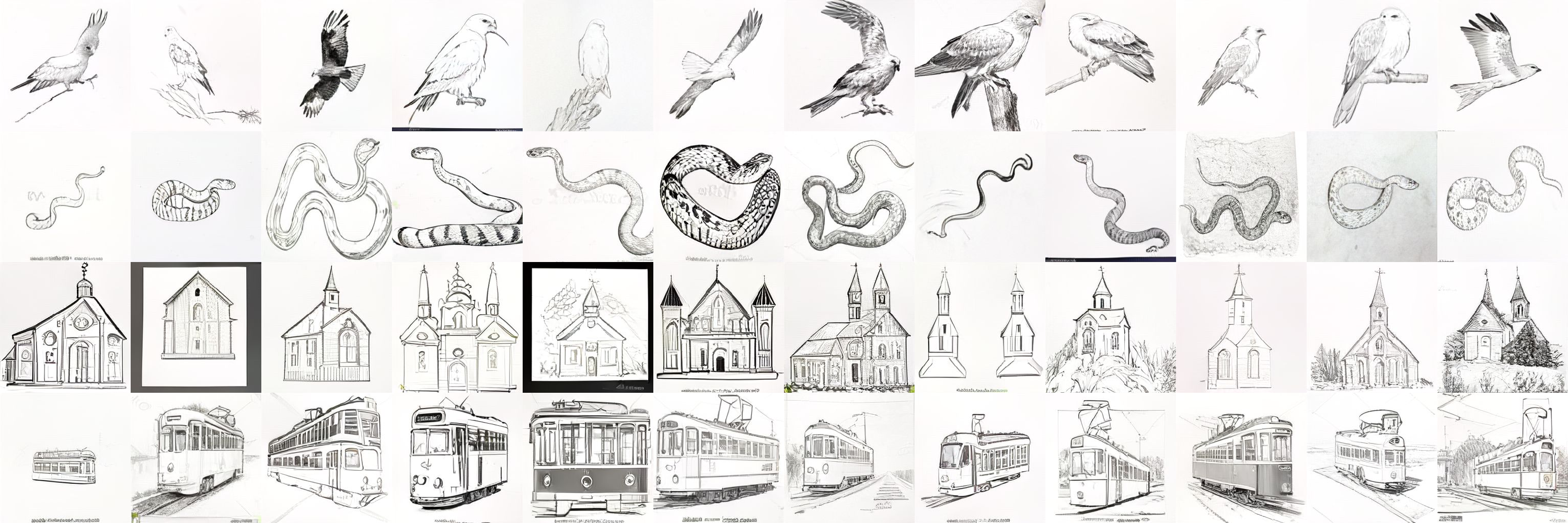}
        \caption{Full (${\approx}\,23$ images per class).}
        \label{fig:app_imagenetr_to_sketch_synthesis_more_full}
    \end{subfigure}
    \caption{Zero-shot domain adaptive image synthesis from ImageNet-R sketch~\cite{hendrycks2021many} with varying amount of training images, (a) 1 or (b) 5 images per class, or (c) full, \ie, approximately 23 images per class, in the target domain. Class conditions are ``kite (\texttt{n01608432})'', ``water snake (\texttt{n01737021})'', ``church (\texttt{n03028079})'', ``streetcar (\texttt{n04335435})'', which do not exist in the ImageNet-R sketch.}
    \label{fig:app_imagenetr_to_sketch_synthesis_more}
\end{figure*}

Another interesting observation is that the generation diversity is affected by the amount of training images. \cref{fig:app_imagenetr_to_sketch_synthesis_more} shows synthesized images by the models trained with $1$ or $5$ images per class or full (${\approx}\,23$ images per class) training data. When comparing the most visually compelling results, we do not see too much differences between these models, and even synthesized images from the model trained with 1 images per class are highly realistic sketch images. However, the model trained with a few training images often sample from very narrow distribution, leading to a reduced diversity. As a result, as confirmed in \cref{tab:imagenetr_to_sketch}, we get worse FIDs. The model performs similarly to the model trained on the full dataset in terms of a zero-shot FID when the number of training images per class is more than or equal to $x$.

\subsection{Extended Ablation Study}
\label{sec:app_exp_ablation}

\subsubsection{Does accurate class prediction matter?}
\label{sec:app_exp_ablation_prediction}

Our method is designed to disentangle semantic information and the target domain information using class-specific and class-agnostic prompts, respectively. As discussed in \cref{sec:exp_zdais_small_training_qualitative_ablation}, our method is able to find reasonable semantic concepts within the vocabulary of the pretrained model. One question arises -- what if our model fails to recognize semantic concept in the target images, or they cannot be described using the source classes? To this end, we conduct additional experiments with abstract painting images as target domain images. 

\cref{tab:class_prediction_supp} shows images used for the target domain and the class prediction by class affinity matrix \textbf{A}. We show results in \cref{fig:zdais_oneshot_training_supp}. To our surprise, our method learns to compose the style of the target domain to a certain degree with semantic knowledge from the source domain. However, we see some failure cases from the third row of \cref{fig:zdais_oneshot_training_supp}. For example, synthesized images conditioned on the class ``convertible'' and ``table lamp'' are seriously entangled with the rotating pattern of the target domain training image. From \cref{tab:class_prediction_supp}, we find predictions being highly uncorrelated. 

\subsection{Zero-shot Domain Adaptation by Synthesis}
\label{sec:app_exp_zsda}

\subsubsection{Generation Details}
\label{sec:app_exp_zsda_generation}

We largely follow the implementation details in \cref{sec:app_exp_zdais_implementation details}. The class-conditional MaskGIT trained on the ImageNet is used as a source model, and we train a class-agnostic prompt of length $S\,{=}\,128$ following \cref{sec:app_exp_zdais_many_training}. The visual prompts are trained on the training set of each domain and split for $200$ epochs.

For synthesis, we condition on the closest ImageNet classes we could find for each class of Office-home dataset. For example, for ``Backpack'' class of Office-home dataset, we choose ``backpack (\texttt{n02769748})'' and ``mailbag (\texttt{n03709823})''. However, for some classes, there is not exactly matching classes between Office-home and ImageNet. To identify the closest classes, we train our model on the \textsc{Real World} domain of Office-home dataset, and match the classes using the class affinity matrix \textbf{A}. This results in matching ``Batteries'' class to ``oil filter'' and ``lighter''. Though they are not exactly matching, we find synthesized images are still useful to train a classifier. We list matching classes between Office-home and ImageNet in \cref{tab:app_officehome_to_imagenet_matching}. \cref{fig:officehome_synthesis} visualizes zero-shot domain adaptive synthesized images. We highlight how our model synthesizes images in the \textsc{Clipart} domain, where the domain gap from the ImageNet might be the largest, by zero-shot transfer.

\subsubsection{Classification Details}
\label{sec:app_exp_zsda_classification}

We mostly follow the same experimental setting and methodology as \cite{jhoo2021collaborative} with some exceptions. Following \cite{jhoo2021collaborative}, we fine-tuned from an imagenet pretrained Resnet-50. We used a jax implementation using the weights from the standard pytorch model zoo model. We did a small hyperparameter search on our source-only classification results and used those hyperparameters all our classification training including Source + Synth training. We found this gave a much stronger source-only transfer baseline and in some cases even beat the domain adapataion results from \cite{jhoo2021collaborative}. We hyperparameter searched over learning rate in $\{0.01, 0.005, 0.001, 0.0005, 0.0001\}$ and selected $lr=0.0005$. We used SGD with nesterov momentum. We found as is commonly known that this has resulted in a model with much better robustness than the ADAM trained model. We follow the learning rate schedule from \cite{jhoo2021collaborative} and cut the learning rate by 0.1 a $\frac{1}{3}$ and $\frac{2}{3}$ into the total training steps. Following \cite{jhoo2021collaborative} we also used a batch size of 8 and trained on one GPU. We chose to train all experiments for a consistent $6250$ steps instead of 50 epochs in \cite{jhoo2021collaborative} since all the datasets have a wildly varying number of examples. Following \cite{jhoo2021collaborative} we reported the best accuracy on the target domain during each training run which we evaluated every $250$ steps.

For our method where the classification model is fine-tuned on the combined source and synthesized images, we used the same hyperparameters for comparability, but added 1000 generated target domain examples from each class. We did a hyperparameter search over how often to sample from the source or the synth examples, $\{25\%, 50\%, 75\%, 100\%\}$ and used $75\%$ of samples chosen were synthesized examples.

\section{Comparison to Other Methods}
\label{sec:app_comparison}

\subsection{Comparison to Cross-Domain Adaptation~\cite{yang2021one}}
\label{sec:app_comparison_to_genda}

Cross-domain adaptation (CDA), briefly studied in \cite{yang2021one}, is relevant to \expset{}. While CDA has been demonstrated on unconditional generation model of single category (\eg, church generation model), our method makes use of class-conditional generation model to transfer the style of the target domain to diverse semantic classes. 

In \cref{fig:supp_compare_with_genda} we make a direct comparison to GenDA~\cite{yang2021one}. Results in the first row are taken from \cite{yang2021one} and those in the second row are synthesized by our method. For synthesis, we condition with the church class (\texttt{n03028079}). When transferring from the ``Superman'' (\cref{fig:supp_compare_with_genda_superman}), GenDA generates yellow-ish church images, likely by learning color from the face region. On the other hand, our method generates cartoon-like church images with more abstract style. This implies that our method not only learns color or tone from the target domain training image, but higher-level concepts like cartoon or comic book.

\subsection{Comparison to Neural Style Transfer}
\label{sec:app_comparison_to_nst}

As we discuss in \cref{sec:related}, \expset{} could be relevant to the neural style transfer (NST)~\cite{gatys2015neural,ghiasi2017exploring,deng2022stytr2} in that both methods aim to generate images that belong to the target domain in some way. The critical difference is that the NST is a translation method, \ie, image-in, image-out, while \expset{} is a class-conditional generation method, \ie., class-in, image-out. As such, NST evaluates structural similarity between input and output images, while we check semantic similarity between input class and output image for our method. Therefore, two methods are not directly comparable using the same evaluation criteria.

Nevertheless, in \cref{fig:app_comparision_to_nst} we collect results from a few state-of-the-art neural style transfer methods~\cite{ghiasi2017exploring,deng2022stytr2} and our method to highlight some qualitative differences. As mentioned, structural or semantic similarities are own characteristics of each method, so should not be counted as a factor for comparison. In terms of transferring the style, both methods seem to work well. We observe that synthesized images by our method are more harmonized than those by NST as there is no structural similarity constraint applied to our method. This is also seen from \cref{fig:officehome_synthesis} or \cref{fig:imagenetr_synthesis} where our model not only transfer artistic style, but also the high-level concepts, such as ``Clipart'', ``Origami'', or ``Toy''. On the other hand, synthesized images by NST usually reflect the style more faithfully.

\section{Information on Images for Target Domain}
\label{sec:app_image_source}

We provide the sources of images used in \cref{tab:supp_image_source} for experiments in \cref{sec:exp_zdais_small_training_qualitative} and \cref{tab:supp_image_source_imagenetr} for experiments in \cref{sec:exp_zdais_small_training_quantitative} for reproducible research.

\begin{minipage}{\linewidth}
    \centering
    
    {
        \captionsetup{type=table}
        \begin{tabular}{c|c|c||c|c|c|c}
        \toprule
            Synthesis & Shot & Step & $S\,{=}\,1$ & $S\,{=}\,16$ & $S\,{=}\,64$ & $S\,{=}\,128$ \\
             \midrule
            Source & -- & 12 & \multicolumn{4}{c}{77.0} \\
            Target, in-dist~\cite{sohn2022visual} & $\approx$23 (full) & 12 & 30.2 & 31.4 & 31.5 & 31.5\\
            \midrule
            Target, zero-shot (ours) & 1 & 12 & 66.1 & 46.1 & 41.3 & 35.8 \\
            Target, zero-shot (ours) & 2 & 12 & 40.2 & 26.6 & 24.8 & 22.1 \\
            Target, zero-shot (ours) & 3 & 12 & 30.1 & 19.9 & 17.6 & 17.2 \\
            Target, zero-shot (ours) & 4 & 12 & 26.5 & 17.8 & 16.6 & 15.7 \\
            Target, zero-shot (ours) & 5 & 12 & 21.5 & 16.1 & \textbf{14.7} & \textbf{14.7} \\
            Target, zero-shot (ours) & $\approx$23 (full) & 12 & 18.9 & 15.1 & \textbf{14.4} & \textbf{14.8} \\
            \midrule
            Target, zero-shot (ours) & 5 & 36 & 20.7 & 15.0 & 13.0 & 12.9 \\
            Target, zero-shot (ours) & $\approx$23 (full) & 36 & 18.4 & 13.7 & 12.6 & \textbf{12.2} \\
        \bottomrule
        \end{tabular}
        \vspace{0.05in}
        \captionof{table}{FID (lower the better) of models trained on ImageNet-R sketch~\cite{hendrycks2021many} and tested on ImageNet-sketch~\cite{wang2019learning}. The number of training images (``Shot'') and the number of decoding step (``Step'') are denoted.}
        \label{tab:imagenetr_to_sketch}
    }
    \vspace{0.1in}
    
    {
        \captionsetup{type=figure}
        \includegraphics[width=0.93\textwidth]{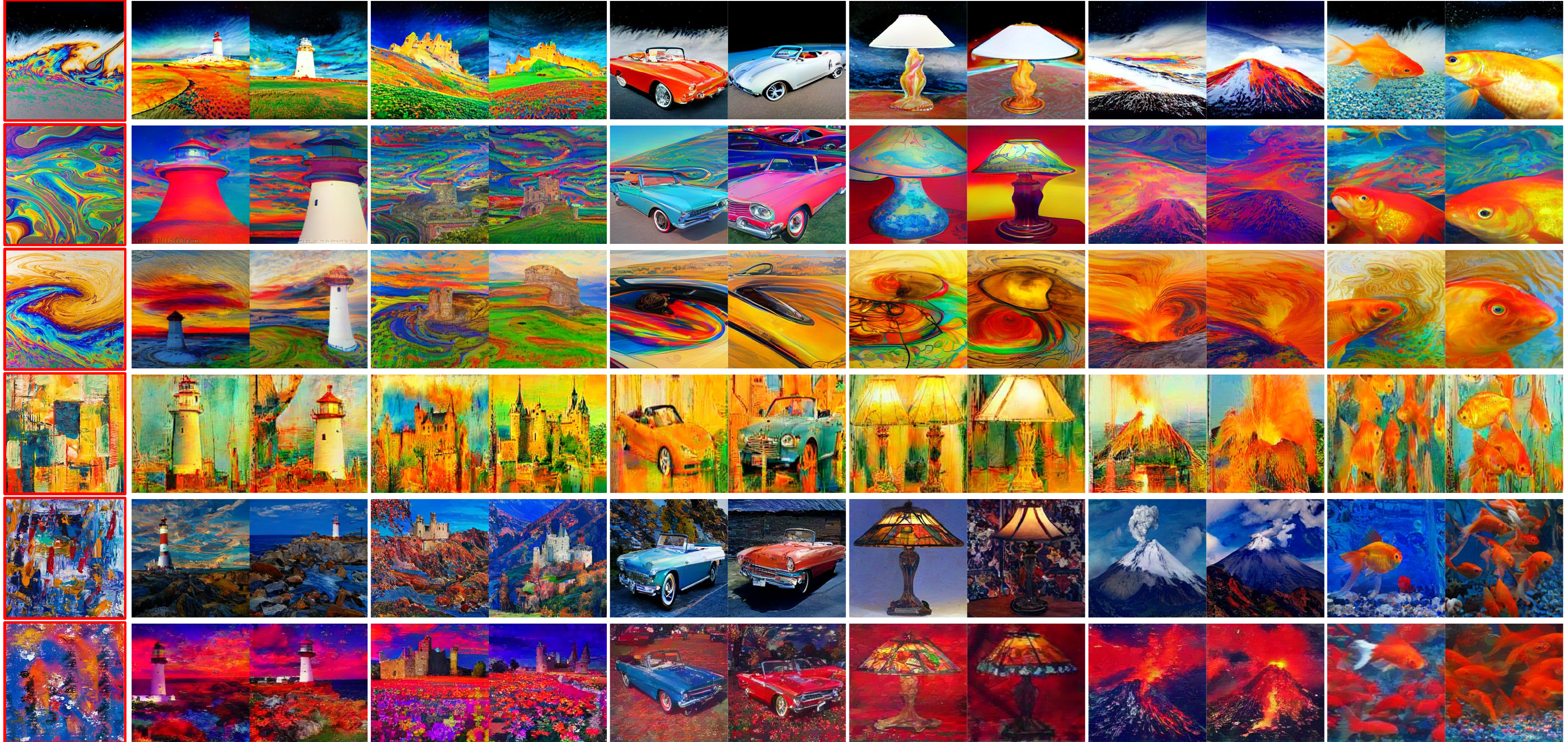}
        \captionof{figure}{Zero-shot domain adaptive image synthesis from a single training image. Images inside red boxes are single training image for each target domain, and the rest are synthesized. Class conditions are ``lighthouse'', ``castle'', ``convertible'', ``table lamp'', ``volcano'' and ``goldfish''. Information on training images are in \cref{sec:app_image_source}.}
        \label{fig:zdais_oneshot_training_supp}
    }
    \vspace{0.1in}
    
    {
        \captionsetup{type=table}
        \resizebox{0.9\linewidth}{!}{%
        \begin{tabular}{c|@{\hspace{0.1in}}c@{\hspace{0.1in}}|c|@{\hspace{0.1in}}c@{\hspace{0.1in}}|c|@{\hspace{0.1in}}c@{\hspace{0.1in}}}
            \toprule
            image & top 5 classes predicted by \textbf{A} &image & top 5 classes predicted by \textbf{A} & image & top 5 classes predicted by \textbf{A} \\
            \midrule
             \parbox[c]{0.5in}{\includegraphics[width=0.5in]{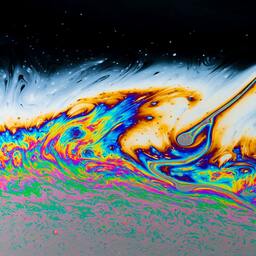}} & \makecell{carousel, skunk feather boa,\\website, borzoi} & \parbox[c]{0.5in}{\includegraphics[width=0.5in]{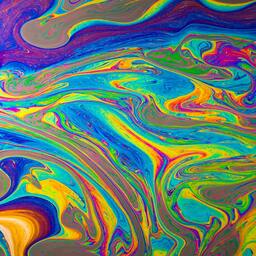}} & \makecell{carousel, cowboy boat, quilt,\\bubble, rock beauty fish} & \parbox[c]{0.5in}{\includegraphics[width=0.5in]{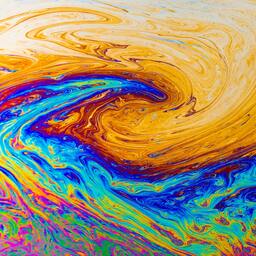}} & \makecell{neck brace, gyromitra, poncho,\\tailed frog, clothes iron} \\
            \midrule
             \parbox[c]{0.5in}{\includegraphics[width=0.5in]{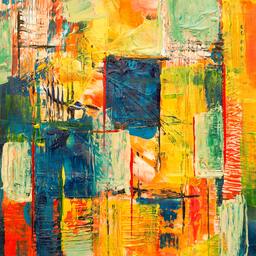}} & \makecell{paintbrush, common gallinule,\\bookstore, stone wall, crate} & \parbox[c]{0.5in}{\includegraphics[width=0.5in]{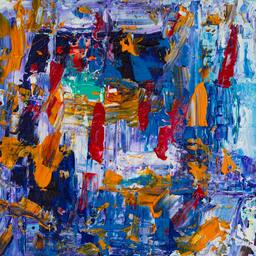}} & \makecell{paintbrush, goldfish, shower\\curtain, snowmobile, fox squirrel} & \parbox[c]{0.5in}{\includegraphics[width=0.5in]{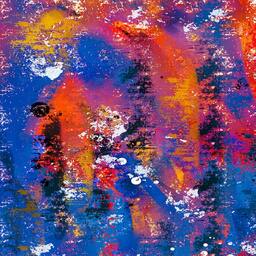}} & \makecell{paintbrush, ski, goldfish,\\feature boa, snorkel} \\
            \bottomrule
        \end{tabular}
        }
        \vspace{0.05in}
        \captionof{table}{Top 5 ImageNet classes by the class affinity matrix \textbf{A}. Relevant classes to the images are bold-faced.}
        \label{tab:class_prediction_supp}
    }
\end{minipage}

\begin{table}[t]
    \centering
    \resizebox{0.8\linewidth}{!}{%
    \begin{tabular}{@{\hspace{0.1in}}l|@{\hspace{0.1in}}l}
    \toprule
        Office-home class & ImageNet class \\
        \midrule
        Alarm Clock & analog clock (\texttt{n02708093}), digital clock (\texttt{n03196217}), stopwatch (\texttt{n04328186}) \\
        Backpack & backpack (\texttt{n02769748}), mailbag (\texttt{n03709823}) \\
        Batteries$^\dagger$ & oil filter (\texttt{n03843555}), lighter (\texttt{n03666591}) \\
        Bed & quilt (\texttt{n04033995}), four-poster (\texttt{n03388549}) \\
        Bike & mountain bike (\texttt{n03792782}), unicycle (\texttt{n04509417}), bicycle-built-for-two (\texttt{n02835271}) \\
        Bottle & pop bottle (\texttt{n03983396}), water bottle (\texttt{n04557648}), beer bottle (\texttt{n02823428}) \\
        Bucket & bucket (\texttt{n02909870}), pot (\texttt{n03991062}), milk can (\texttt{n03764736}), caldron (\texttt{n02939185}) \\
        Calculator & remote control (\texttt{n04074963}), hand-held computer (\texttt{n03485407}) \\
        Calendar$^\dagger$ & menu (\texttt{n07565083}) \\
        Candles & candle (\texttt{n02948072}), torch (\texttt{n04456115}) \\
        Chair & rocking chair (\texttt{n04099969}), throne (\texttt{n04429376}), barber chair (\texttt{n02791124}) \\
        Clipboards$^\dagger$ & menu (\texttt{n07565083}) \\
        Computer & desktop computer (\texttt{n03180011}) \\
        Couch & studio couch (\texttt{n04344873}) \\
        Curtains & window shade (\texttt{n04590129}), shower curtain (\texttt{n04209239}) \\
        Desk Lamp & table lamp (\texttt{n04380533}) \\
        Drill & power drill (\texttt{n03995372}) \\
        Eraser & rubber eraser (\texttt{n04116512}) \\
        Exit Sign & street sign (\texttt{n06794110}) \\
        Fan & electric fan (\texttt{n03271574}) \\
        File Cabinet & file (\texttt{n03337140}) \\
        Flipflops & sandal (\texttt{n04133789}) \\
        Flowers & daisy (\texttt{n11939491}), yellow lady's slipper (\texttt{n12057211}) \\
        Folder & envelope (\texttt{n03291819}) \\
        Fork$^\dagger$ & spatula (\texttt{n04270147}) \\
        Glasses & sunglasses (\texttt{n04356056}) \\
        Hammer & hammer (\texttt{n03481172}) \\
        Helmet & crash helmet (\texttt{n03127747}), football helmet (\texttt{n03379051}) \\
        Kettle & teapot (\texttt{n04398044}) \\
        Keyboard & computer keyboard (\texttt{n03085013}), space bar (\texttt{n04264628}), typewriter keyboard (\texttt{n04505470}) \\
        Knives & scabbard (\texttt{n04141327}), cleaver (\texttt{n03041632}), letter opener (\texttt{n03658185}) \\
        Lamp Shade & lampshade (\texttt{n03637318}) \\
        Laptop & laptop (\texttt{n03642806}), notebook (\texttt{n03832673}) \\
        Marker & ballpoint (\texttt{n02783161}) \\
        Monitor & screen (\texttt{n04152593}), monitor (\texttt{n03782006}) \\
        Mop & swab (\texttt{n04367480}), broom (\texttt{n02906734}) \\
        Mouse & mouse (\texttt{n03793489}) \\
        Mug & coffee mug (\texttt{n03063599}), cup (\texttt{n07930864}) \\
        Notebook & binder (\texttt{n02840245}) \\
        Oven & rotisserie (\texttt{n04111531}), microwave (\texttt{n03761084}) \\
        Pan & frying pan (\texttt{n03400231}), tray (\texttt{n04476259}), dutch oven (\texttt{n03259280}) \\
        Paper Clip & safety pin (\texttt{n04127249}) \\
        Pen & ballpoint (\texttt{n02783161}), fountain pen (\texttt{n03388183}) \\
        Pencil & pencil sharpener (\texttt{n03908714}) \\
        Postit Notes$^\dagger$ & envelope (\texttt{n03291819}) \\
        Printer & printer (\texttt{n04004767}) \\
        Push Pin$^\dagger$ & pinwheel (\texttt{n03944341}), syringe (\texttt{n04376876}) \\
        Radio & radio (\texttt{n04041544}) \\
        Refrigerator & refrigerator (\texttt{n04070727}) \\
        Ruler & rule (\texttt{n04118776}) \\
        Scissors$^\dagger$ & screwdriver (\texttt{n04154565}), letter opener (\texttt{n03658185}), can opener (\texttt{n02951585}), corkscrew (\texttt{n03109150}) \\
        Screwdriver & screwdriver (\texttt{n04154565}) \\
        Shelf & plate rack (\texttt{n03961711}) \\
        Sink & washbasin (\texttt{n04553703}) \\
        Sneakers & running shoe (\texttt{n04120489}) \\
        Soda & pop bottle (\texttt{n03983396}) \\
        Speaker & loudspeaker (\texttt{n03691459}) \\
        Spoon & ladle (\texttt{n03633091}), wooden spoon (\texttt{n04597913}) \\
        TV & television (\texttt{n04404412}) \\
        Table & dining table (\texttt{n03201208}) \\
        Telephone & dial telephone (\texttt{n03187595}), pay-phone (\texttt{n03902125}) \\
        ToothBrush$^\dagger$ & paintbrush (\texttt{n03876231}) \\
        Toys$^\dagger$ & toyshop (\texttt{n04462240}), maraca (\texttt{n03720891}) \\
        Trash Can & ashcan (\texttt{n02747177}) \\
        Webcam$^\dagger$ & projector (\texttt{n04009552}), tripod (\texttt{n04485082}) \\
    \bottomrule
    \end{tabular}
    }
    \vspace{0.05in}
    \caption{Class matching between Office-home~\cite{venkateswara2017deep} and ImageNet~\cite{deng2009imagenet}. Office-home classes without exact match are denoted with $\dagger$.}
    \label{tab:app_officehome_to_imagenet_matching}
\end{table}

\end{document}